\documentclass[10pt,journal,compsoc]{IEEEtran}

\usepackage[breaklinks=true,colorlinks,citecolor=black,linkcolor=black,urlcolor=black]{hyperref}
\usepackage{url}            
\usepackage{booktabs}       
\usepackage{amsfonts}       
\usepackage{nicefrac}       
\usepackage{microtype}      
\usepackage{xcolor}         
\usepackage{subfigure}
\usepackage{times}
\usepackage{graphicx}
\usepackage{amsmath}
\usepackage{amssymb}
\usepackage{colortbl}

\usepackage{color}
\usepackage{xcolor}
\usepackage{multirow}
\usepackage{overpic}

\usepackage[nocompress]{cite} 
\usepackage{ragged2e}

\usepackage{cancel}  
\usepackage[normalem]{ulem}

\graphicspath{{./Imgs/}}

\usepackage{pifont}
\newcommand{\cmark}{\ding{51}}%
\newcommand{\xmark}{\text{\ding{55}}}

\usepackage{multirow}

\usepackage{makecell}

\usepackage{silence}
\def\etal{{\em et al.}}
\def\aka{\emph{a.k.a.,~}}

\def\sArt{state-of-the-art~}
\def\ie{\emph{i.e.,~}}
\def\eg{\emph{e.g.,~}}
\def\etc{{\em etc}}
\def\LKA{large kernel attention}
\def\VAN{Visual Attention Network}

\newcommand{\figref}[1]{Fig.~\ref{#1}}
\newcommand{\tabref}[1]{Tab.~\ref{#1}}
\newcommand{\secref}[1]{Sec.~\ref{#1}}

\newcommand{\myPara}[1]{\vspace{.1in}\noindent \textbf{#1}\quad }

\begin{document}

\title{Visual Attention Network}

\author{Meng-Hao Guo, Cheng-Ze Lu, Zheng-Ning Liu, 
  Ming-Ming Cheng and Shi-Min Hu
\IEEEcompsocitemizethanks{\IEEEcompsocthanksitem M.-H. Guo and 
  S.-M. Hu are with the Department of Computer Science, 
  Tsinghua University, Beijing, China. 
  Emails:gmh20@mails.tsinghua.edu.cn, shimin@tsinghua.edu.cn.
  Shi-Min Hu is the corresponding author.
\IEEEcompsocthanksitem C.-Z. Lu and M.-M. Cheng are with 
  Nankai University University, Tianjin, China.
  Emails: czlu919@outlook.com, cmm@nankai.edu.cn.
\IEEEcompsocthanksitem Z.-N. Liu are with 
  Fitten Tech, Beijing, China.
  Emails: lzhengning@gmail.com.
}

\thanks{Manuscript received April 19, 2005; revised August 26, 2015.}}

\markboth{Journal of \LaTeX\ Class Files,~Vol.~14, No.~8, August~2015}%
{Shell \MakeLowercase{\textit{et al.}}: Bare Demo of IEEEtran.cls for Computer Society Journals}

\IEEEtitleabstractindextext{%
\begin{abstract} 
  While originally designed for natural language processing tasks,
  the self-attention mechanism
  has recently taken various computer vision areas by storm.
  However, the 2D nature of images brings three challenges for 
  applying self-attention in computer vision.
  (1) Treating images as 1D sequences neglects their 2D structures.
  (2) The quadratic complexity is too expensive for high-resolution images.
  (3) It only captures spatial adaptability but ignores channel adaptability.
  In this paper, we propose a novel
  linear attention named 
  \LKA~(LKA) to enable 
  self-adaptive and long-range correlations in self-attention
  while avoiding its shortcomings.
  Furthermore, we present a neural network based on LKA, 
  namely Visual Attention Network (VAN).
  While extremely simple, 
  VAN surpasses similar size vision transformers(ViTs) and 
  convolutional neural networks(CNNs) in various tasks, 
  including image classification, object detection, semantic segmentation,
  panoptic segmentation,
  pose estimation, \etc.
  For example, VAN-B6 achieves 87.8\% accuracy on ImageNet benchmark
  and set new state-of-the-art performance (58.2 PQ) for panoptic segmentation.
  Besides, VAN-B2 surpasses Swin-T
  4\% mIoU (50.1 vs. 46.1) for semantic segmentation on ADE20K benchmark,
  2.6\% AP (48.8 vs. 46.2)  for object detection on COCO dataset.
  It provides a novel method and a simple yet strong baseline for 
  the community.
  Code is available at
  \url{https://github.com/Visual-Attention-Network}.
\end{abstract}

}

\maketitle

\IEEEdisplaynontitleabstractindextext
\IEEEpeerreviewmaketitle

\IEEEraisesectionheading{\section{Introduction}\label{sec:introduction}}

\IEEEPARstart{A}{s} the basic feature extractor, vision backbone
is a fundamental research topic in the computer vision field.
Due to remarkable feature extraction performance,
convolutional neural networks (CNNs)
\cite{lecun1998gradient,lecun1989backpropagation,krizhevsky2012imagenet} 
are indispensable topic in the last decade.
After the AlexNet~\cite{krizhevsky2012imagenet} reopened the 
deep learning decade,
a number of breakthroughs have been made to get more powerful vision backbones,
by using deeper network~\cite{simonyan2014very,he2016deep},
more efficient architecture
\cite{howard2017mobilenets,xie2017aggregated,zhang2018shufflenet},
stronger multi-scale ability
\cite{huang2017densely,szegedy2015going,pami21Res2net}, 
and attention mechanisms~\cite{hu2018squeeze,dosovitskiy2020image}.
Due to translation invariance property
and shared sliding-window strategy~\cite{sermanet2013overfeat}, 
CNNs are inherently efficient for various vision tasks 
with arbitrary sized input.
More advanced vision backbone networks often results in
significant performance gain in various tasks, including 
image classification~\cite{he2016deep,dosovitskiy2020image,liu2021swin}, 
object detection~\cite{dai2017deformable}, 
semantic segmentation~\cite{xie2021segformer} and 
pose estimation~\cite{wang2020deep}.

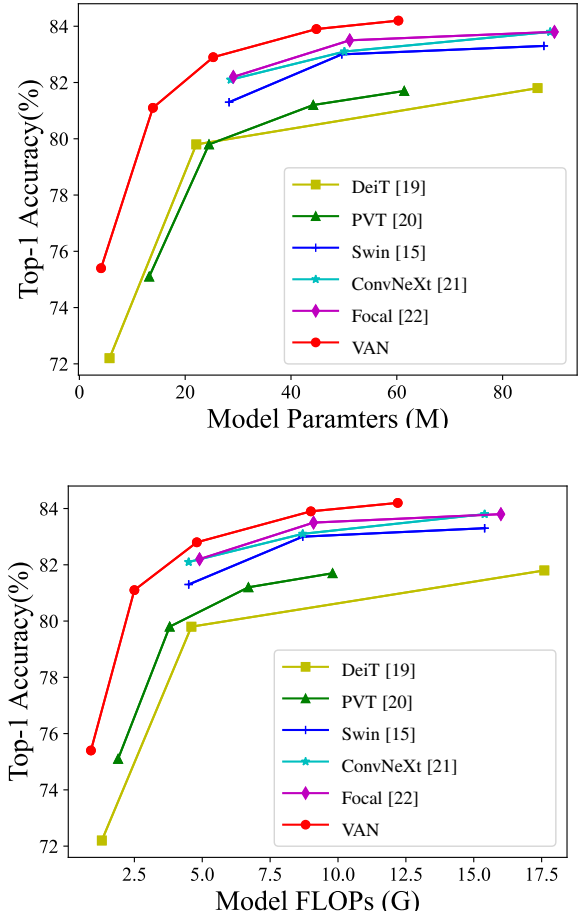
\begin{figure}[t]
  \centering
  \scriptsize
  \begin{overpic}[width=.47\textwidth]{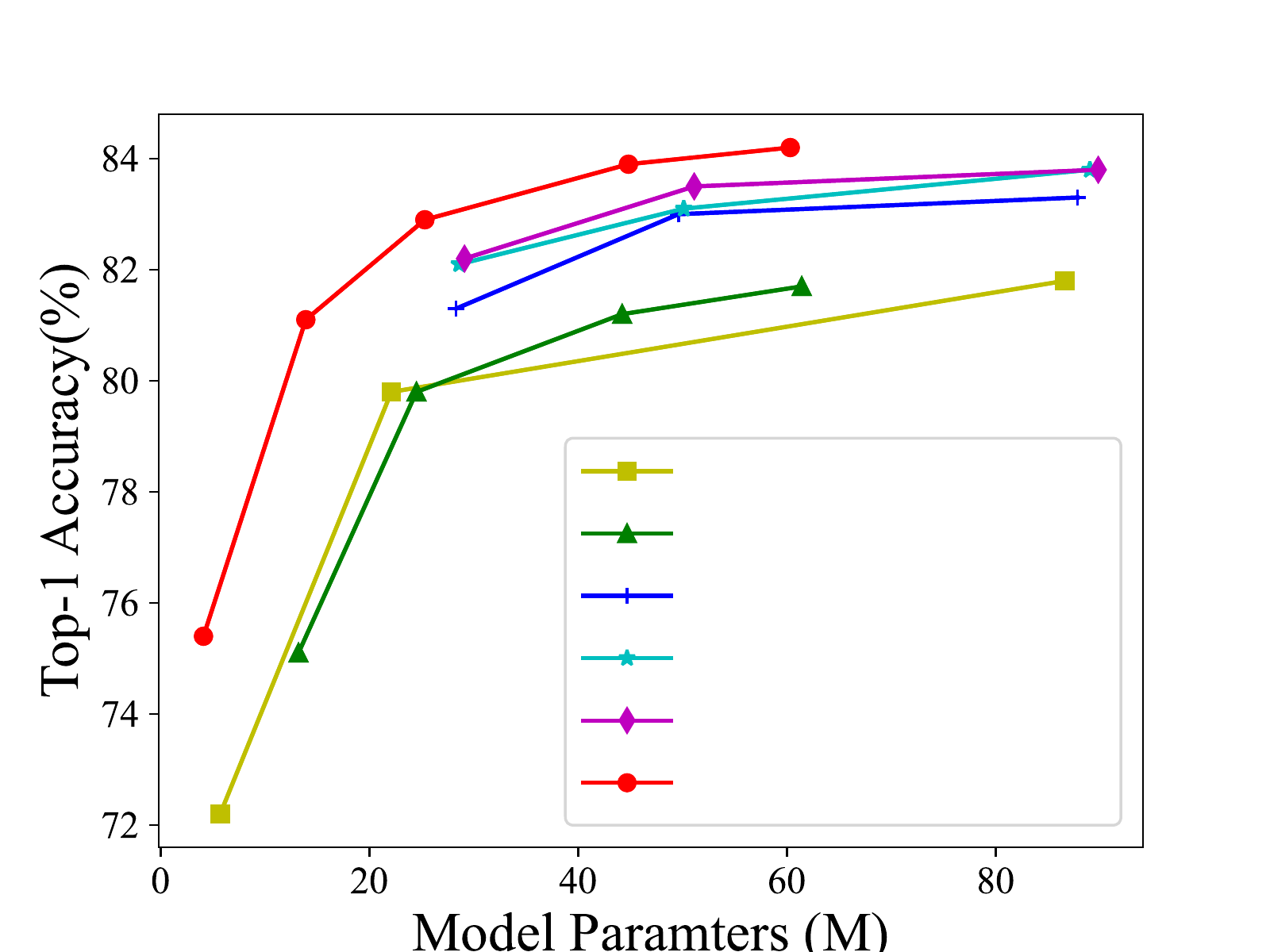}
    \put(55, 36.5){DeiT \cite{touvron2021training}}
    \put(55, 31.5){PVT \cite{wang2021pyramid}}
    \put(55, 26.5){Swin \cite{liu2021swin}}
    \put(55, 21.5){ConvNeXt \cite{liu2022convnet}}
    \put(55, 16.5){Focal \cite{yang2021focal}}
    \put(55, 11.5){VAN}
  \end{overpic} \hspace{2mm}
  \begin{overpic}[width=.47\textwidth]{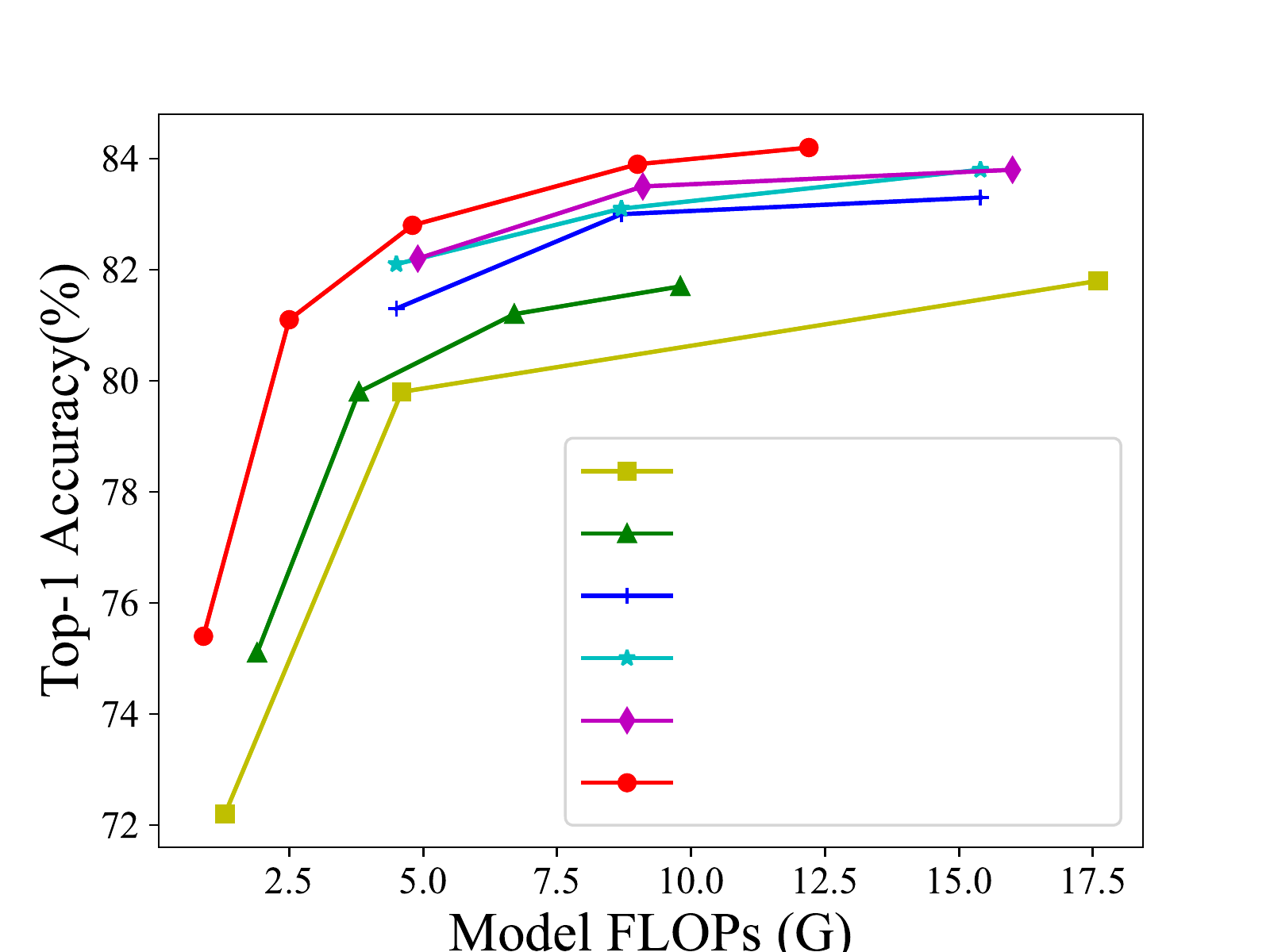}
    \put(55, 36.5){DeiT \cite{touvron2021training}}
    \put(55, 31.5){PVT \cite{wang2021pyramid}}
    \put(55, 26.5){Swin \cite{liu2021swin}}
    \put(55, 21.5){ConvNeXt \cite{liu2022convnet}}
    \put(55, 16.5){Focal \cite{yang2021focal}}
    \put(55, 11.5){VAN}
  \end{overpic} \hspace{2mm}
  \caption{Results of different models on ImageNet-1K validation set. 
  Comparing the performance of recent models 
  DeiT~\cite{touvron2021training}, PVT~\cite{wang2021pyramid}, 
  Swin Transformer~\cite{liu2021swin}, ConvNeXt~\cite{liu2022convnet}, 
  Focal Transformer~\cite{yang2021focal} and our VAN.
  Above: Accuracy-Parameters trade-off diagram.
  Under: Accuracy-FLOPs trade-off diagram.
}\label{fig:results}
\end{figure}

Based on observed reaction times and estimated signal transmission times 
along biological pathways \cite{gottlieb1998representation}, 
cognitive psychology \cite{treisman1980feature}
and neuroscience \cite{wolfe2004attributes} researchers believe that 
human vision system processes only parts of possible stimuli in detail, 
while leaving the rest nearly unprocessed.
Selective attention is an important mechanism for dealing with 
the combinatorial aspects of complex search in vision
\cite{tsotsos1995modeling}.
Attention mechanism can be regarded as an adaptive selecting process 
based on the input feature.
Since the fully attention network \cite{vaswani2017attention} been proposed,
self-attention models (\aka Transformer) quickly becomes 
the dominated architecture~\cite{devlin2018bert,brown2020language}
in natural language processing (NLP).
Recently, Dosovitskiy~\etal~\cite{dosovitskiy2020image} 
propose the vision transformer (ViT),
which introduces transformer backbone into computer vision 
and outperforms well-known CNNs on image classification tasks.
Benefited from its powerful modeling capabilities, 
transformer-based vision backbones quickly occupy the leaderboards of 
various tasks, 
including object detection~\cite{liu2021swin}, 
semantic segmentation~\cite{xie2021segformer}, \etc.

Even with remarkable success, 
convolution operation and self-attention still have their shortcomings. 
Convolution operation adopts static weight and lacks adaptability, 
which has been proven critical~\cite{hu2018squeeze,dai2017deformable}.
As originally designed for 1D NLP tasks,
self-attention~\cite{dosovitskiy2020image,dosovitskiy2020image} 
regards 2D images as 1D sequences, 
which destroys the crucial 2D structure of the image. 
It is also difficult to process high-resolution images due to its 
quadratic computational and memory overhead.
Besides, self-attention is a special attention 
that only considers the adaptability in spatial dimension 
but ignores the adaptability in channel dimension, 
which is also important for visual tasks
\cite{hu2018squeeze,woo2018cbam,wang2020ecanet,el2021xcit}.

In this paper, we propose a novel linear
attention mechanism dubbed \LKA~(LKA),
which is tailored for visual tasks.
LKA absorbs the advantages of convolution and self-attention, 
including local structure information, long-range dependence, and adaptability. 
Meanwhile, it avoids their disadvantages such as ignoring adaptability 
in channel dimension.
Based on the LKA, we present a novel vision backbone called \VAN~(VAN) 
that significantly surpasses well-known CNN-based and transformer-based backbones.
The contributions of this paper are summarized as follows:
\begin{itemize}
  \item We design a novel linear 
    attention mechanism named LKA for computer vision, 
    which considers the pros of both convolution and self-attention,
    while avoiding their cons.
    Based on LKA, we further introduce a 
    simple vision backbone called VAN.
  \item We show that VANs outperform the similar level ViTs and CNNs 
    in extensive experiments on various tasks, 
    including image classification, object detection, 
    semantic segmentation, instance segmentation, pose estimation, \etc.
\end{itemize}

\section{Related Work}

\subsection{Convolutional Neural Networks}

How to effectively compute powerful feature representations is the 
most fundamental problem in computer vision.
Convolutional neural networks (CNNs) 
\cite{lecun1998gradient,lecun1989backpropagation},
utilize local contextual information and 
translation invariance properties to greatly 
improve the effectiveness of neural networks.
CNNs quickly become the mainstream framework in computer vision 
since AlexNet~\cite{krizhevsky2012imagenet}.
To further improve the usability, 
researchers put lots of effort in making the CNNs 
deeper \cite{simonyan2014very,he2016deep,huang2017densely,han2021demystifying,bello2021revisiting,szegedy2015going},
and lighter 
\cite{howard2017mobilenets,sandler2018mobilenetv2,zhang2018shufflenet}. 
Our work has similarity with MobileNet~\cite{howard2017mobilenets}, 
which decouples a standard convolution into two parts, 
a depthwise convolution and a pointwise convolution
(\aka $1\times 1$ Conv \cite{2014Network}).
Our method decomposes a convolution into three parts: 
depthwise convolution, 
depthwise and dilated convolution~\cite{chen2014semantic,yu2015multi}, 
and pointwise convolution. 
Benefiting from this decomposition, 
our method is more suitable for 
efficiently decomposing large kernel convolutions.
We also introduce attention mechanism into our method to 
obtain adaptive property.

\subsection{Visual Attention Methods}\label{sec.related_attention}

Attention mechanism can be regarded as an adaptive selection process 
according to the input feature, 
which is introduced into computer vision in RAM~\cite{mnih2014recurrent}. 
It has provided benefits in many visual tasks, 
such as image classification~\cite{hu2018squeeze,woo2018cbam}, 
object detection~\cite{dai2017deformable,hu2018relation} and 
semantic segmentation~\cite{yuan2020object,geng2021attention}. 
Attention in computer vision can be divided into 
four basic categories~\cite{guo2021attention_survey}, 
including channel attention, spatial attention, 
temporal attention and branch attention, 
and their combinations such as channel \& spatial attention.  
Each kind of attention has a different effect in visual tasks.

Originating from NLP~\cite{vaswani2017attention,devlin2018bert},
self-attention is a special kind of attention mechanism.
Due to its effectiveness of capturing the long-range dependence 
and adaptability, 
it is playing an increasingly important role in computer vision
\cite{wang2018non,fu2019dual,ramachandran2019stand,Bello_AANet,yuan2018ocnet,zhang2019self,xie2018attentional}.
Various deep self-attention networks (\aka vision transformers)
\cite{dosovitskiy2020image,carion2020end,liu2021swin,guo_pct,srinivas2021bottleneck,wang2021pyramid,Yuan_2021_ICCV,liu2021decoupled,bello2021lambdanetworks,xu2021vitae,liu2021fuseformer,bao2021beit,liu2022dabdetr,wu2021cvt,liu2021query2label,wu2021p2t,he2021masked} 
have achieved significantly better performance than the mainstream CNNs 
on different visual tasks, 
showing the huge potential of attention-based models. 
However, self-attention is originally designed for NLP.
It has three shortcomings when dealing with computer vision tasks. 
(1) It treats images as 1D sequences which neglects the 2D structure of images.
(2) The quadratic complexity is too expensive for high-resolution images.
(3) It only achieves spatial adaptability but ignores the adaptability 
in channel dimension. 
For vision tasks, different channels often represent different objects
\cite{chen2017sca,guo2021attention_survey}. 
Channel adaptability is also proven important for visual tasks
\cite{hu2018squeeze,woo2018cbam,qin2021fcanet,wang2020ecanet,chen2017sca}. 
To solve these problems,  
we propose a novel visual attention method, namely, LKA.
It involves the pros of self-attention such as adaptability and 
long-range dependence.
Besides, it benefits from the advantages of convolution such as 
making use of local contextual information.

\subsection{Vision MLPs}

Multilayer Perceptrons (MLPs)
\cite{rosenblatt1958perceptron,rumelhart1985learning} 
were a popular tool for computer vision before CNNs appearing. 
However, due to high computational requirements and low efficiency,
the capability of MLPs was been limited in a long time.
Some recent research successfully decouple standard MLP into 
spatial MLP and channel MLP 
\cite{tolstikhin2021mlp,guo2021beyond,touvron2021resmlp,liu2021pay}.
Such decomposition allows significant computational cost and parameters
reduction,
which release the amazing performance of MLP.
Readers are referred to recent surveys \cite{guo2021can,liu2021ready} 
for a more comprehensive review of MLPs.
The most related MLP to our method is the gMLP~\cite{liu2021pay}, 
which not only decomposes the standard MLP but also involves 
the attention mechanism.
However, gMLP has two drawbacks. 
On the one hand, 
gMLP is sensitive to input size and can only process fixed-size images.
On the other hand, gMLP only considers the global information of the image 
and ignore their local structure.
Our method can make full use of its advantages and avoid its shortcomings.

\begin{figure}[t]
  \centering 
  \footnotesize
  \begin{overpic}[width=\linewidth]{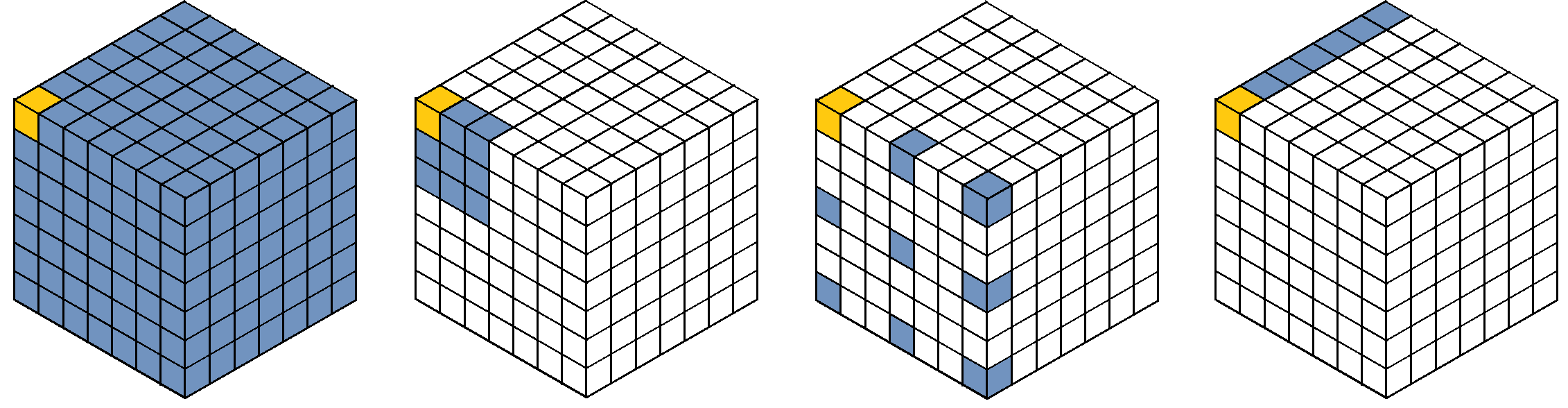}
    \put(-3,12){$H$}
    \put(23.1,12){$=$}\put(48.8,12){$+$}\put(74.3,12){$+$}
    \put(3.4,0){$W$} 
    \put(17,0.8){$C$}
  \end{overpic}
\caption{Decomposition diagram of large-kernel convolution. 
  A standard convolution can be decomposed into three parts: 
  a depth-wise convolution (DW-Conv), 
  a depth-wise dilation convolution (DW-D-Conv), 
  and a pointwise convolution (1$\times$1 Conv). 
  The colored grids represent the location of convolution kernel and the 
  yellow grid means the center point. 
  The diagram shows that a 13$\times$13 convolution is decomposed into 
  a 5$\times$5 depth-wise convolution, 
  a 5$\times$5 depth-wise dilation convolution with dilation rate 3, 
  and a pointwise convolution. 
  Note: zero paddings are omitted in the above figure.
}\label{fig:decomposition}
\end{figure}

\section{Method}

\begin{figure*}[t]   
    \centering
    \small
    \begin{overpic}[width=\linewidth]{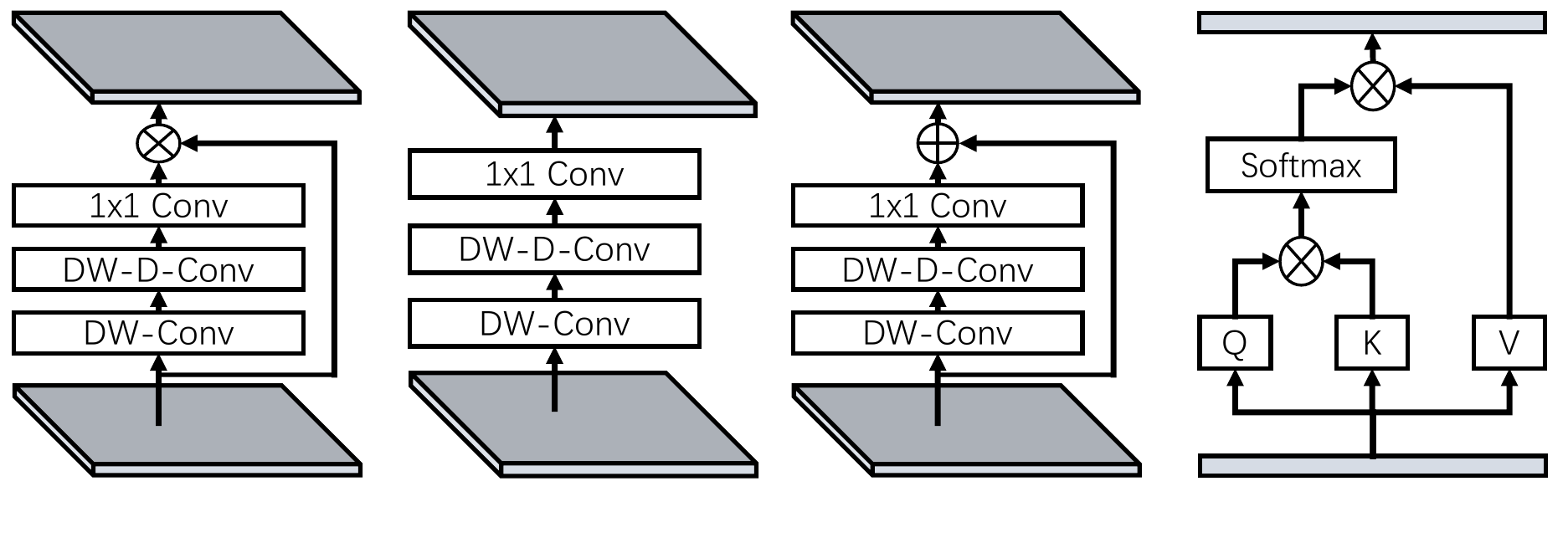}
      \put(9,0){(a) LKA}
      \put(30,0){(b) Non-Attention}
      \put(54,0){(c) Non-Attention(Add)}
      \put(80,0){(d) Self-Attention}
    \end{overpic}\hspace{1pt}
    \caption{The structure of different modules:
      (a) the proposed Large Kernel Attention (LKA); 
      (b) non-attention module;
      (c) replace multiplication in LKA with addition ;
      (d) self-attention. It is worth noting that (d) is designed for 1D sequences.
    }\label{fig:attention}
\end{figure*}

\begin{figure}[t]
  \centering 
  \footnotesize
  \begin{overpic}[width=0.9\linewidth]{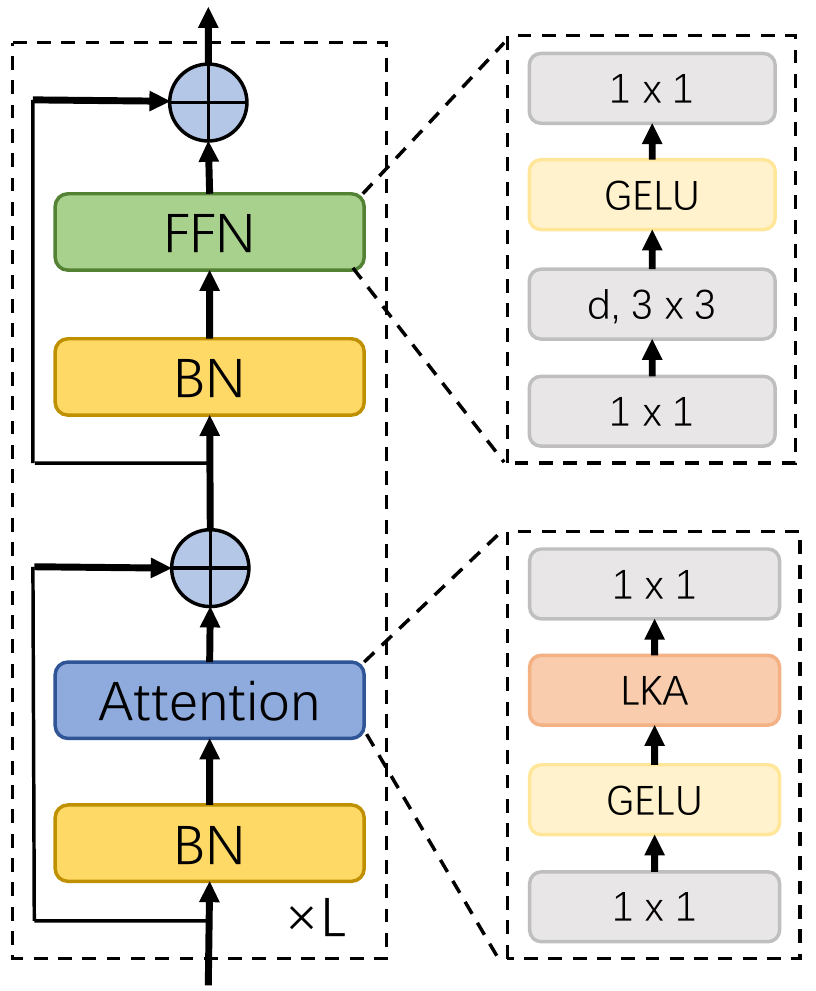}
  \end{overpic}
\caption{A stage of VAN. d means depth wise convolution. k $\times$ k denotes k $\times$ k convolution.
}\label{fig:vanstage}
\end{figure}

\subsection{Large Kernel Attention} \label{sec.attention}

Attention mechanism can be regarded as an adaptive selection process, 
which can select the discriminative features and 
automatically ignore noisy responses according to the input features.
The key step of attention mechanism is producing attention map 
which indicates the importance of different parts.
To do so,
we should learn the relationship between different features. 

There are two well-known methods to build relationship between different 
parts.
The first one is adopting self-attention mechanism
\cite{wang2018non,yuan2018ocnet,zhang2019self,dosovitskiy2020image} 
to capture long-range dependence. 
There are three obvious shortcomings for self-attention applied in computer vision
which have been listed in \secref{sec.related_attention}. 
The second one is to use large kernel convolution
\cite{woo2018cbam,wang2017residual,hu2018gather,park2018bam} 
to build relevance and produce attention map. 
There are still obvious cons in this way. 
Large-kernel convolution brings a huge amount of computational overhead 
and parameters.

To overcome above listed cons and make use of the pros of self-attention 
and large kernel convolution, 
we propose to decompose a large kernel convolution operation 
to capture long-range relationship.
As shown in \figref{fig:decomposition}, 
a large kernel convolution can be divided into three components: 
a spatial local convolution (depth-wise convolution), 
a spatial long-range convolution (depth-wise dilation convolution), and 
a channel convolution (1$\times$1 convolution). 
Specifically.
we can decompose a $K \times K$ convolution into 
a $\lceil \frac{K}{d} \rceil \times \lceil \frac{K}{d} \rceil$ depth-wise dilation convolution with dilation $d$, 
a $(2d-1) \times (2d-1)$ depth-wise convolution 
and a 1$\times$1 convolution.
%
Through the above decomposition, 
we can capture long-range relationship with slight computational cost 
and parameters. 
After obtaining long-range relationship, 
we can estimate the importance of a point and generate attention map.
As demonstrated in \figref{fig:attention}(a), 
the LKA module can be written as
\begin{align}
  Attention &= \text{Conv}_{1\times1}\text{(DW-D-Conv(DW-Conv(F)))}, \\
  Output &= Attention \otimes F.
\end{align}
Here, $F \in \mathbb{R}^{C \times H \times W}$ is the input feature. 
$Attention \in \mathbb{R}^{C \times H \times W}$ denotes attention map.
The value in attention map indicates the importance of each feature.  
\textbf{$\otimes$} means element-wise product.
Different from common attention methods, 
LKA dose not require an additional normalization function 
like sigmoid and softmax, which is demonstrated in~\tabref{tab_ablation}.
We also believe 
the key characteristics of attention methods 
is adaptively adjusting output based on input feature, 
but not the normalized attention map.
As shown in \tabref{Tab.operation_properties}, 
our proposed LKA combines the advantages of convolution and self-attention. 
It takes the local contextual information, large receptive field, linear complexity
and dynamic process into consideration. 
Furthermore, LKA not only achieves the adaptability in the spatial dimension 
but also the adaptability in the channel dimension.
It worth noting that different channels often represent different objects 
in deep neural networks~\cite{guo2021attention_survey,chen2017sca} 
and adaptability in the channel dimension is also important for visual tasks.

\newcommand{\cmplx}[1]{$\mathcal{O}(#1)$}

\begin{table}[t]
  \centering
  \setlength{\tabcolsep}{2.0mm}
  \caption{Desirable properties belonging to convolution, self-attention and LKA.}
  \begin{tabular}{l|c|c|c}
      \hline
      \textbf{Properties} & \textbf{Convolution} & \textbf{Self-Attention} & \textbf{LKA} \\
      \hline
      Local Receptive Field  & \cmark & \xmark &  \cmark   \\
      Long-range Dependence &  \xmark & \cmark &  \cmark   \\
      Spatial Adaptability & \xmark & \cmark &  \cmark   \\
      Channel Adaptability & \xmark & \xmark &  \cmark   \\
      \hline
      Computational complexity&\cmplx{n}&\cmplx{n^2}&\cmplx{n} \\
     \hline
  \end{tabular}
  \label{Tab.operation_properties}
\end{table}

\newcommand{\MCols}[2]{\multicolumn{#1}{c|}{#2}}
\newcommand{\TRows}[1]{\multirow{2}{*}{#1}}

\subsection{Visual Attention Network (VAN)} \label{sec.van}

Our VAN has a simple hierarchical structure,
\ie a sequence of four stages with decreasing output spatial resolution,
\ie $\frac{H}{4} \times \frac{W}{4}$, 
$\frac{H}{8} \times \frac{W}{8}$, 
$\frac{H}{16} \times \frac{W}{16}$ and 
$\frac{H}{32} \times \frac{W}{32}$ respectively.
Here, $H$ and $W$ denote the height and width of the input image. 
With the decreasing of resolution, 
the number of output channels is increasing. 
The change of output channel $C_{i}$ is presented in 
\tabref{tab.architecture}.

For each stage as shown in \figref{fig:vanstage},
we firstly downsample the input and use the stride number to control 
the downsample rate.
After the downsample, all other layers in a stage stay the same
output size, \ie spatial resolution and the number of channels.
Then, $L$ groups of batch normalization~\cite{ioffe2015batch}, 
$1\times 1$ Conv, GELU activation~\cite{hendrycks2020gaussian}, \LKA~and 
feed-forward network (FFN) \cite{wang2021pvtv2}
are stacked in sequence to extract features. 
%
%
We design seven architectures VAN-B0, VAN-B1, VAN-B2, VAN-B3, VAN-B4, VAN-B5, VAN-B6
according to the parameters and computational cost.
The details of the whole network are shown in \tabref{tab.architecture}.

\myPara{Complexity analysis.}
We present the parameters and floating point operations (FLOPs)
of our decomposition.
Bias is omitted in the computation process for simplifying format.
We assume that the input and output features have same size 
$H \times W \times C$.
The number of parameters $P(K, d)$ and FLOPs $F(K, d)$ can be denoted as:
\begin{align}\label{eq_params}
  P(K, d) &= C (\lceil \frac{K}{d} \rceil ^ 2  \times C + (2d - 1)^2) + C^2, \\
  F(K, d) &= P(K, d) \times H \times W.
\end{align}
Here, $d$ means dilation rate and $K$ is kernel size. 
According to the formula of FLOPs and parameters, 
the ratio of budget saving is the same for FLOPs and parameters.

\newcommand{\sUp}[1]{}

\begin{table}[t]
  \footnotesize
  \centering
  \setlength{\tabcolsep}{4.4mm}
  \caption{Number of parameters for different forms of a $21 \times 21$ 
    convolution. 
    For instance, when the number of channels $C=32$,
    standard convolution and MobileNet decomposition use $133 \times$
    and $4.5 \times$ more parameters than our decomposition respectively.
  }
  \begin{tabular}{l|c|c|c}\hline
      & Standard &\multicolumn{2}{c}{Decomposition Type} \\ \cline{3-4}
      & Convolution & MobileNet \cite{howard2017mobilenets} & Ours \\ \hline
      C=32  & 451,584\sUp{133}    & 15,136\sUp{4.5} & 3,392   \\
      C=64  & 1,806,336\sUp{205}  & 32,320\sUp{3.7} & 8,832   \\
      C=128 & 7,225,344\sUp{279}  & 72,832\sUp{2.8} & 25,856  \\
      C=256 & 28,901,376\sUp{342} & 178,432\sUp{2.1}& 84,480  \\
      C=512 & 115,605,504\sUp{385}& 487,936\sUp{1.6}& 300,032 \\ \hline
  \end{tabular}
  \label{Tab.operation_params}
\end{table}

\begin{table}[t]
  \centering
  \caption{Ablation study of different modules in LKA.
    Top-1 accuracy (Acc) on ImageNet validation set suggest that 
    each part is critical.
    w/o Attention means we adopt \figref{fig:attention}(b).
  }
  \label{tab_ablation}
  \begin{tabular}{l|c|c|c} \hline
    VAN-B0      & Params. (M) & FLOPs(G)  &  Acc(\%)  \\  \hline
    w/o DW-Conv           & 4.1 & 0.9  &  74.9 \\
    w/o DW-D-Conv         & 4.0 & 0.9 &  74.1 \\
    w/o Attention         & 4.1 & 0.9 &  74.3 \\
    w/o Attention (Add)   & 4.1 & 0.9 &  74.6 \\
    w/o 1 $\times$ 1 Conv & 3.8  & 0.8 &  74.6 \\
    w/ Sigmoid & 4.1  & 0.9  &  75.2 \\
    VAN-B0                   &  4.1 & 0.9 &  75.4 \\ \hline
  \end{tabular}
\end{table}

\begin{table}[t]
  \centering
  \caption{Throughput of Swin transformer and VAN on RTX 3090.}
  \label{tab_speed}
  \begin{tabular}{l|c|c|c} \hline
    Method  & FLOPs(G)  &  \makecell{Throughput  (Imgs/s)} &   Acc(\%)  \\  \hline
    Swin-T   & 4.5 & 821 &  81.3 \\
    Swin-S   & 8.7 & 500 &  83.0 \\
    Swin-B   & 15.4 & 376 &  83.5 \\
    \hline
    VAN-B0  & 0.9 & 2140 &  75.4 \\
    VAN-B1  & 2.5 & 1420 &  81.1 \\
    VAN-B2  & 5.0 & 762 &  82.8 \\
    VAN-B3  & 9.0 & 452 &  83.9 \\
    VAN-B4 & 12.2 & 341 &  84.2 \\
\hline
  \end{tabular}
\end{table}

\begin{figure}[t]
  \centering
  \scriptsize
  \begin{overpic}[width=.47\textwidth]{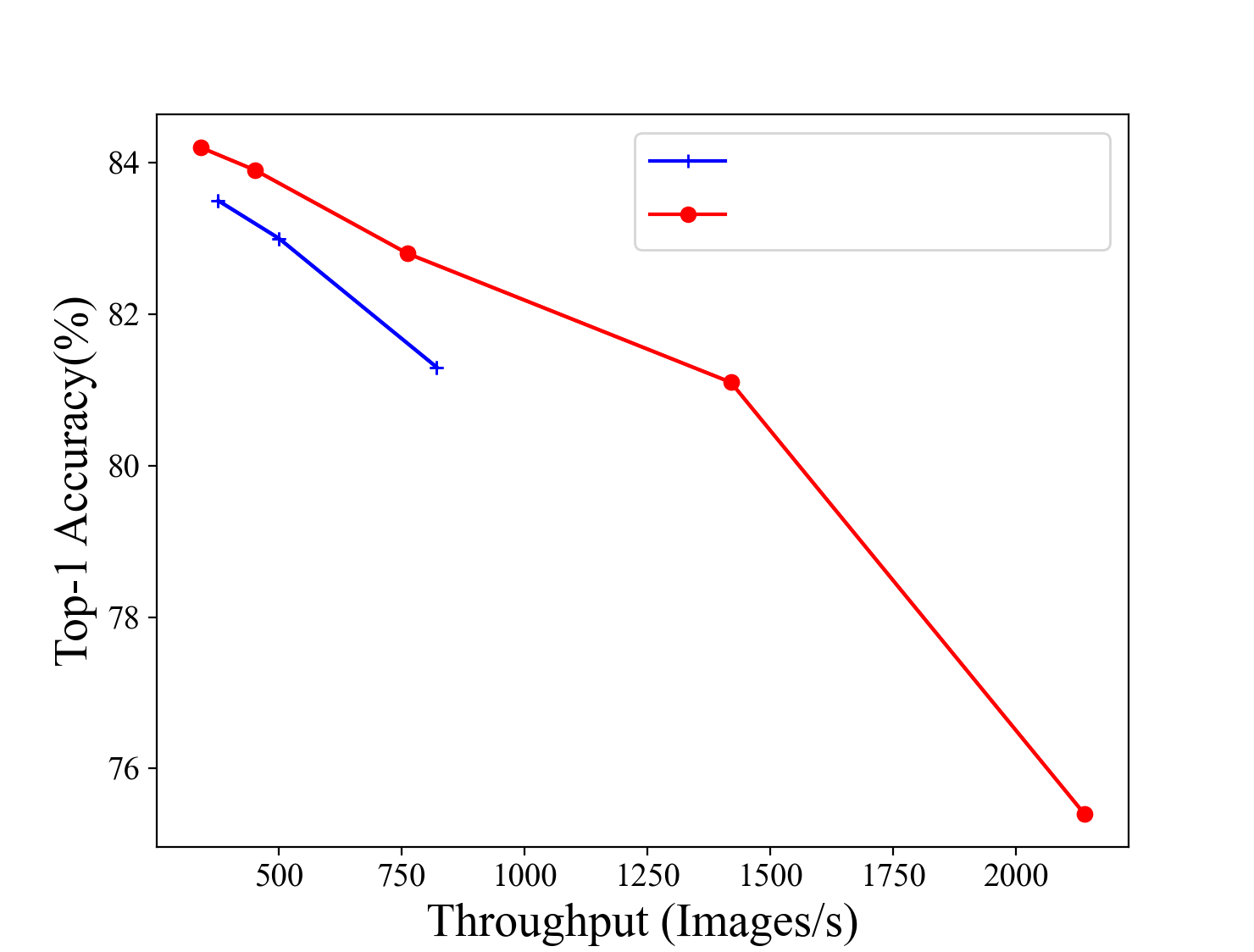}
    \put(61, 62){Swin \cite{liu2021swin}}
    \put(61, 58){VAN}
  \end{overpic}
    \caption{Accuracy-Throughput Diagram. It claerly shows that 
    VAN achieves a better trade-off than swin transformer~\cite{liu2021swin}.
}\label{fig:fps_result}
\end{figure}


\myPara{Implementation details.}
We adopt K = 21 by default.
For $K=21$, the Equ. \eqref{eq_params} takes the minimum value
when $d = 3$, 
which corresponds to $5 \times 5$ depth-wise convolution and 
$7 \times 7$ depth-wise convolution with dilation $3$.
For different number of channels, we show the specific parameters in 
\tabref{Tab.operation_params}. 
It shows that our decomposition owns significant advantages in 
decomposing large kernel convolution in terms of parameters and FLOPs.


\begin{table*}[t]
  \centering
  \setlength{\tabcolsep}{8.0pt}
  \caption{The detailed setting for different versions of the VAN.
  e.r. represents expansion ratio in the feed-forward network.
  }\label{tab.architecture}
  \begin{tabular}{c|c|c|c|c|c|c|c|c|c}
    \hline \hline
    \TRows{stage} & \TRows{output size} & \TRows{e.r.} & 
    \multicolumn{7}{c}{VAN-} \\ \cline{4-10}
     &  & & B0 & B1 & B2 & B3 & B4 & B5 & B6 \\ \hline
    1 & $\frac{H}{4}\times \frac{W}{4} \times C$ & 8 & 
    \makecell{$C=32$ \\ $L=3$} & \makecell{$C=64$ \\ $L=2$} & \makecell{$C=64$ \\ $L=3$} & \makecell{$C=64$ \\ $L=3$} & \makecell{$C=64$ \\ $L=3$} & \makecell{$C=96$ \\ $L=3$}  & \makecell{$C=96$ \\ $L=6$}  \\ \hline
    2 & $\frac{H}{8}\times \frac{W}{8} \times C$ & 8 & 
    \makecell{$C=64$ \\ $L=3$} & \makecell{$C=128$ \\ $L=2$} & \makecell{$C=128$ \\ $L=3$} &  \makecell{$C=128$ \\ $L=5$} & \makecell{$C=128$ \\ $L=6$} &  \makecell{$C=192$ \\ $L=3$} & \makecell{$C=192$ \\ $L=6$}  \\ \hline
    3 & $\frac{H}{16}\times \frac{W}{16} \times C$ & 4 & 
    \makecell{$C=160$ \\ $L=5$} & \makecell{$C=320$ \\ $L=4$} & \makecell{$C=320$ \\ $L=12$} &  \makecell{$C=320$ \\ $L=27$} & \makecell{$C=320$ \\ $L=40$} &  \makecell{$C=480$ \\ $L=24$} & \makecell{$C=384$ \\ $L=90$}  \\ \hline
    4 & $\frac{H}{32}\times \frac{W}{32} \times C$ & 4 & 
    \makecell{$C=256$ \\ $L=2$} & \makecell{$C=512$ \\ $L=2$} & \makecell{$C=512$ \\ $L=3$} &  \makecell{$C=512$ \\ $L=3$} & \makecell{$C=512$ \\ $L=3$} 
     &  \makecell{$C=768$ \\ $L=3$} & \makecell{$C=768$ \\ $L=6$} \\ \hline
    \MCols{3}{Parameters (M)} & 4.1 & 13.9 & 26.6 & 44.8 & 60.3 & 90.0 & 200  \\ \hline
    \MCols{3}{{FLOPs} (G)}    & 0.9 & 2.5  & 5.0 & 9.0 & 12.2 & 17.2 &  38.4 \\ \hline
    \hline
  \end{tabular}
\end{table*}

\section{Experiments}

In this section, quantitative and qualitative experiments are exhibited to
demonstrate the effectiveness and efficiency of the proposed VAN.
We conduct quantitative experiments on ImageNet-1K~\cite{deng2009imagenet} 
and ImageNet-22K image classification dataset,
COCO~\cite{lin2014microsoft} benchmark for object detection, instance segmentation,
panoptic segmentation and pose estimation, 
and ADE20K~\cite{zhou2019semantic} semantic segmentation dataset. 
Furthermore, we visualize the experimental results and 
class activation mapping(CAM)~\cite{zhou2016learning} by using 
Grad-CAM~\cite{selvaraju2017grad} on ImageNet validation set.
Experiments are based on Pytorch~\cite{paszke2019pytorch} and Jittor~\cite{hu2020jittor}.

\subsection{Image Classification}

\subsubsection{ImageNet-1K Experiments}

\myPara{Settings.} We conduct image classification on 
ImageNet-1K~\cite{deng2009imagenet}  dataset. 
It contains 1.28M training images and 50K validation images from 
1,000 different categories.
The whole training scheme mostly follows ~\cite{touvron2021training}.  
We adopt random clipping, random horizontal flipping, 
label-smoothing~\cite{muller2019does}, mixup~\cite{zhang2017mixup}, 
cutmix~\cite{yun2019cutmix} and random erasing~\cite{zhong2020random} 
to augment the training data.
In the training process, 
we train our VAN for 300 epochs by using AdamW
\cite{kingma2017adam,loshchilov2019decoupled} optimizer
with momentum$=$0.9, weight decay=$5\times10^{-2}$ and batch size = 1,024. 
Cosine schedule~\cite{loshchilov2016sgdr} and warm-up strategy are employed 
to adjust the learning rate(LR). 
The initial LR is set to $5\times10^{-4}$. 
We adopt a variant of LayerScale~\cite{touvron2021going} in attention layer
which replaces 
$x_{out} = x + diag(\lambda_{1},\lambda_{2},...,\lambda_{d}) f(x)$
with
$x_{out} = x + diag(\lambda_{1},\lambda_{2},...,\lambda_{d}) (f(x) + x)$ 
with initial value 0.01 and 
achieves a better performance than original LayerScale.
Exponential moving average (EMA)~\cite{polyak1992acceleration} is also 
applied to improve training process.
During the eval stage, 
we report the top-1 accuracy on ImageNet validation set under 
single crop setting.

\myPara{Ablation Study.}
We conduct an ablation study to prove that each component of LKA is critical.
In order to obtain experimental results quickly, 
we choose VAN-B0 as our baseline model.
The experimental results in the \tabref{tab_ablation}~indicate that 
all components in LKA are indispensable to improve performance.

\begin{figure*}[t]   
    \centering
    \small
    \begin{overpic}
    [width=0.3\linewidth]{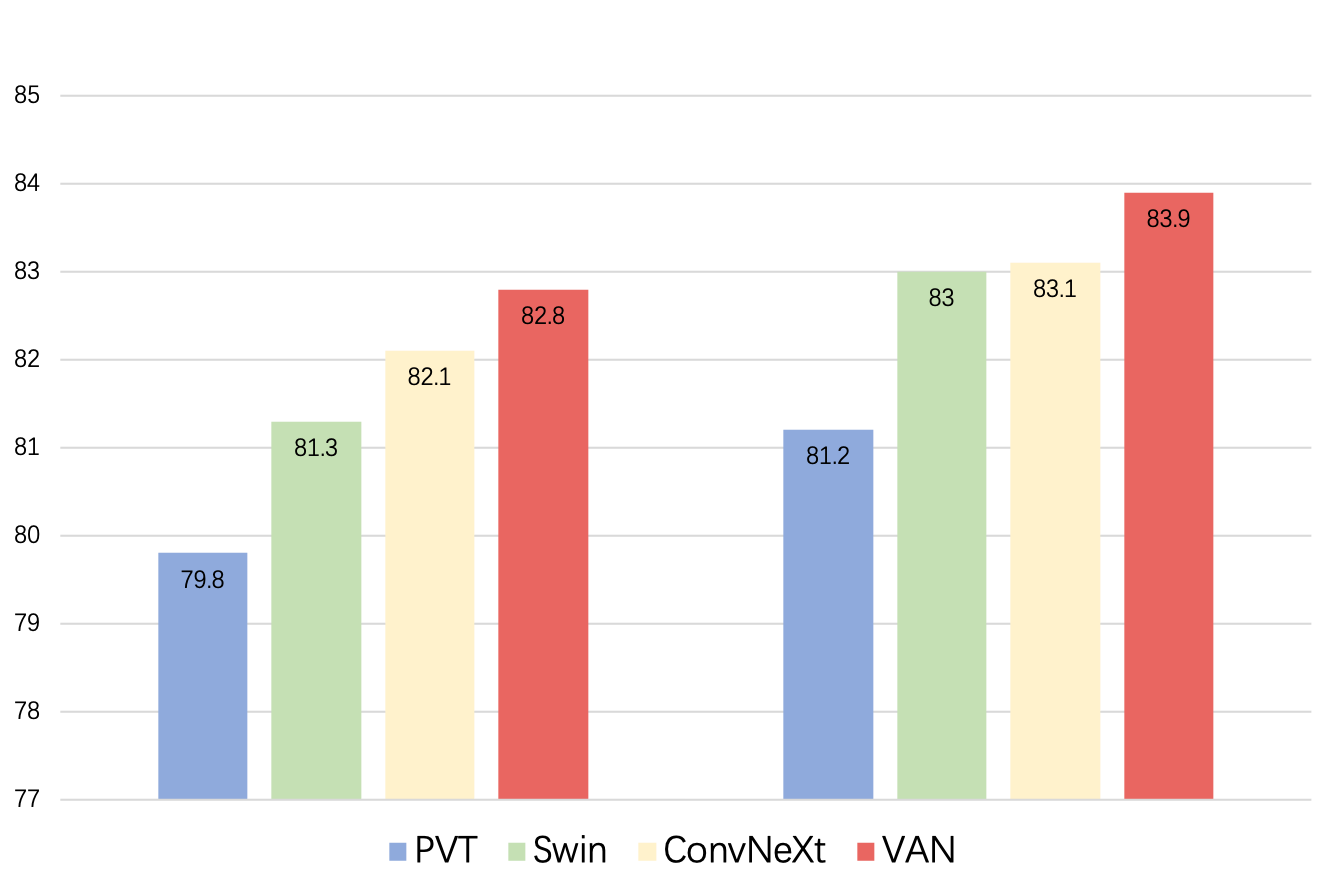}
    \put(5,62){Classification (IN1K)}
    \end{overpic}\hspace{1pt}
    \begin{overpic}
    [width=0.3\linewidth]{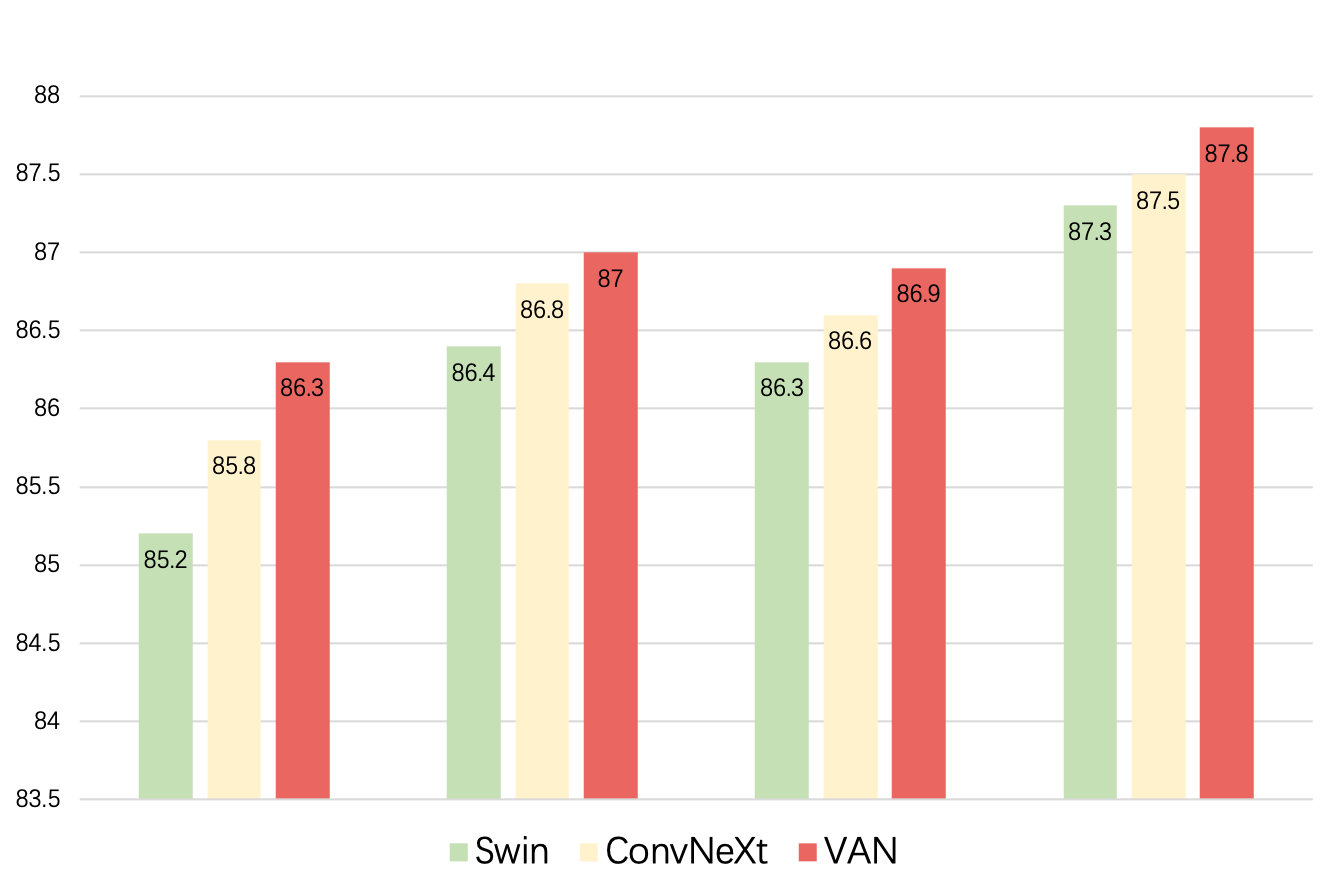}
    \put(5,62){Classification(IN22K pre-training)}
    \end{overpic}\hspace{1pt}
    \begin{overpic}
    [width=0.3\linewidth]{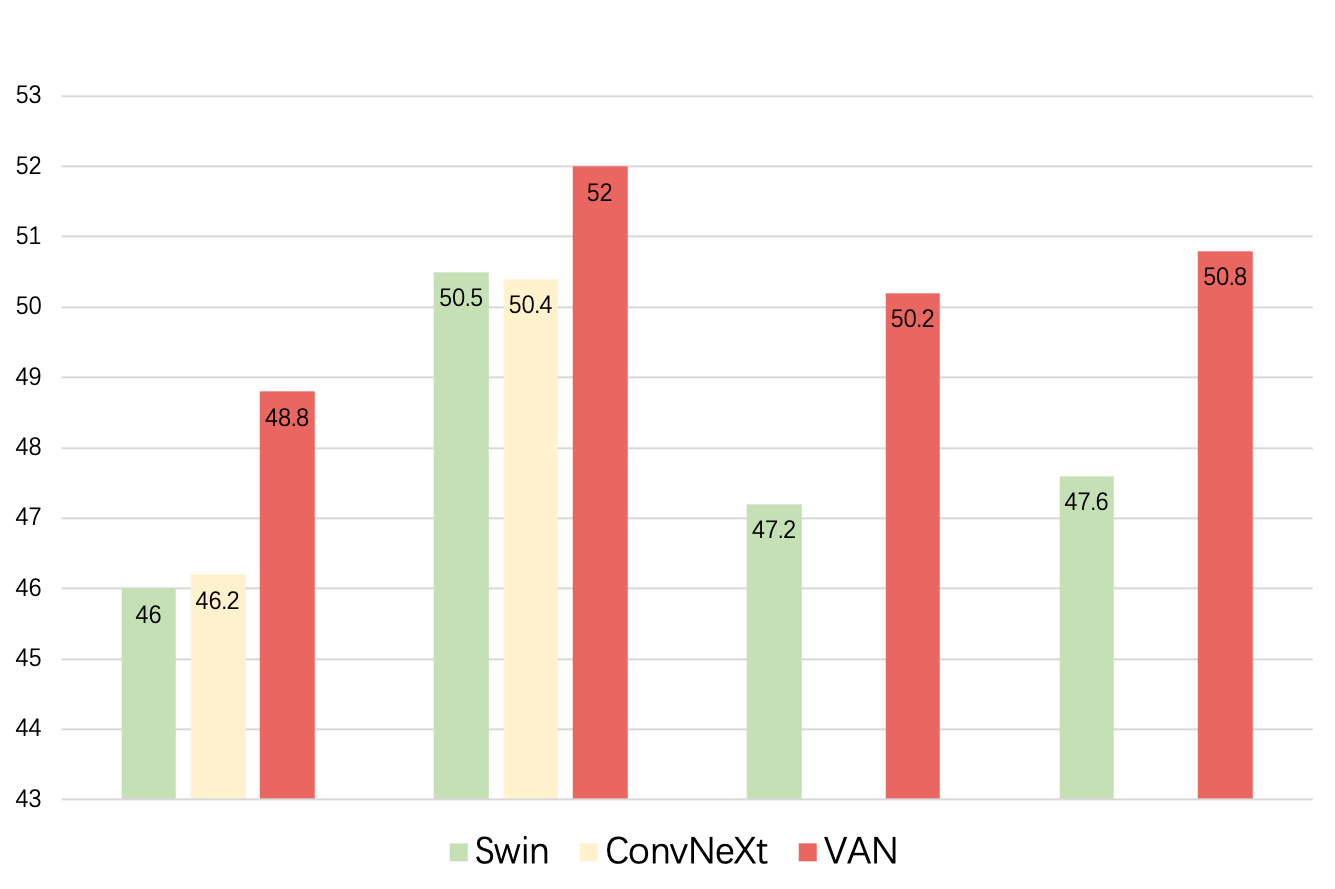}
    \put(5,62){Object Detection}
    \end{overpic}\hspace{1pt} 
    \\
    \begin{overpic}
    [width=0.3\linewidth]{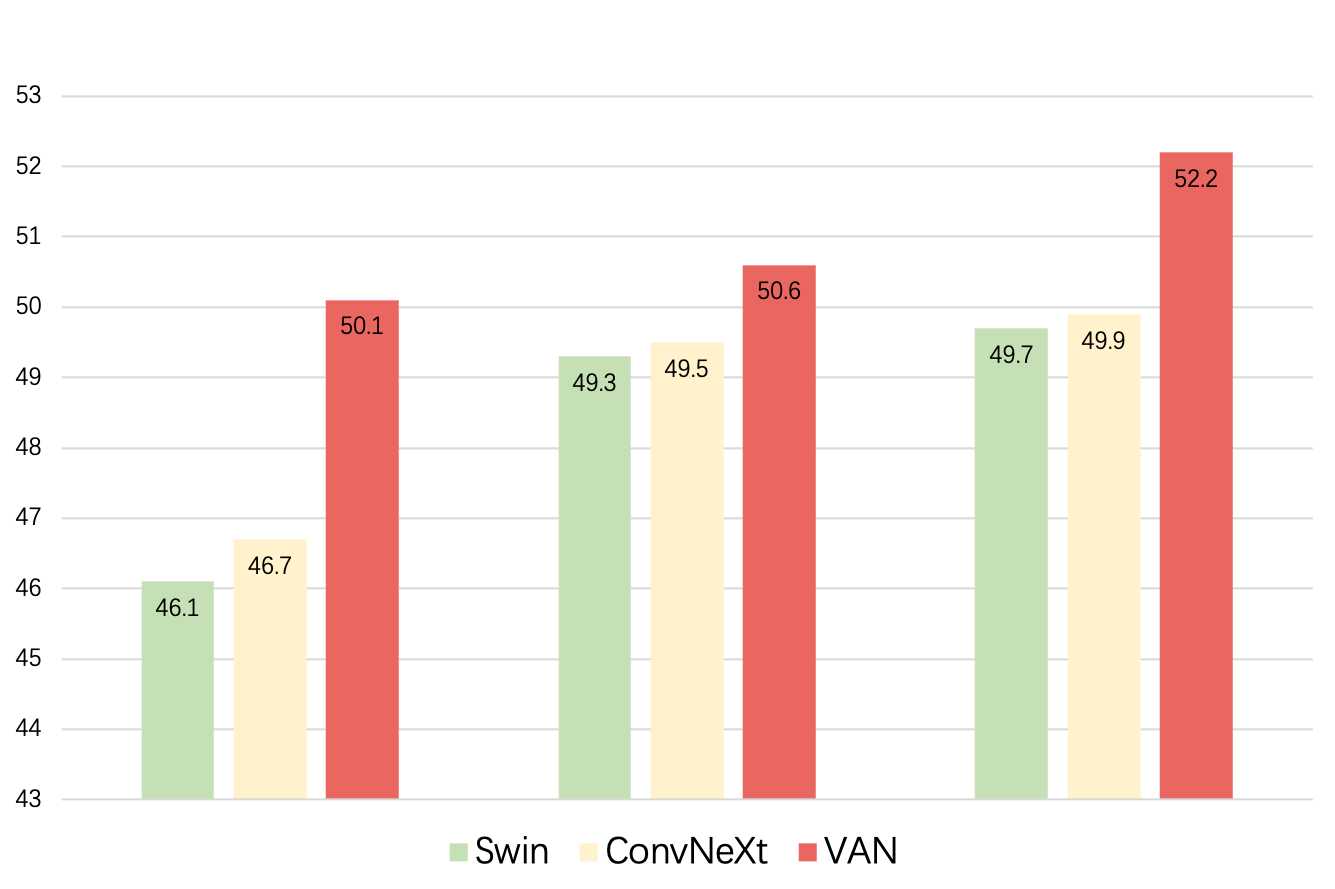}
    \put(5,62){Semantic Segmentation}
    \end{overpic}\hspace{1pt}
    \begin{overpic}
    [width=0.3\linewidth]{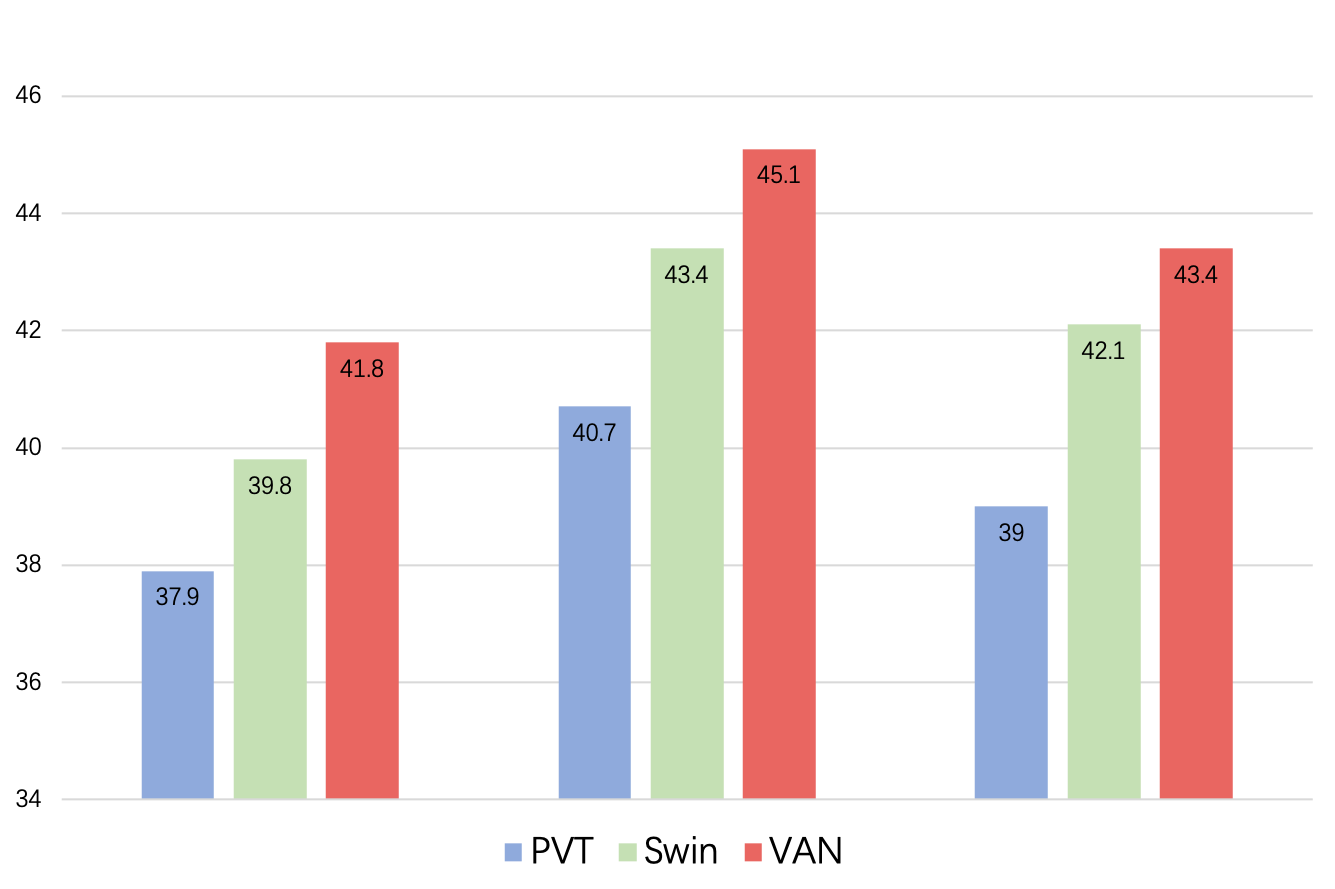}
    \put(5,62){Instance Segmentation}
    \end{overpic}\hspace{1pt}
    \begin{overpic}
    [width=0.3\linewidth]{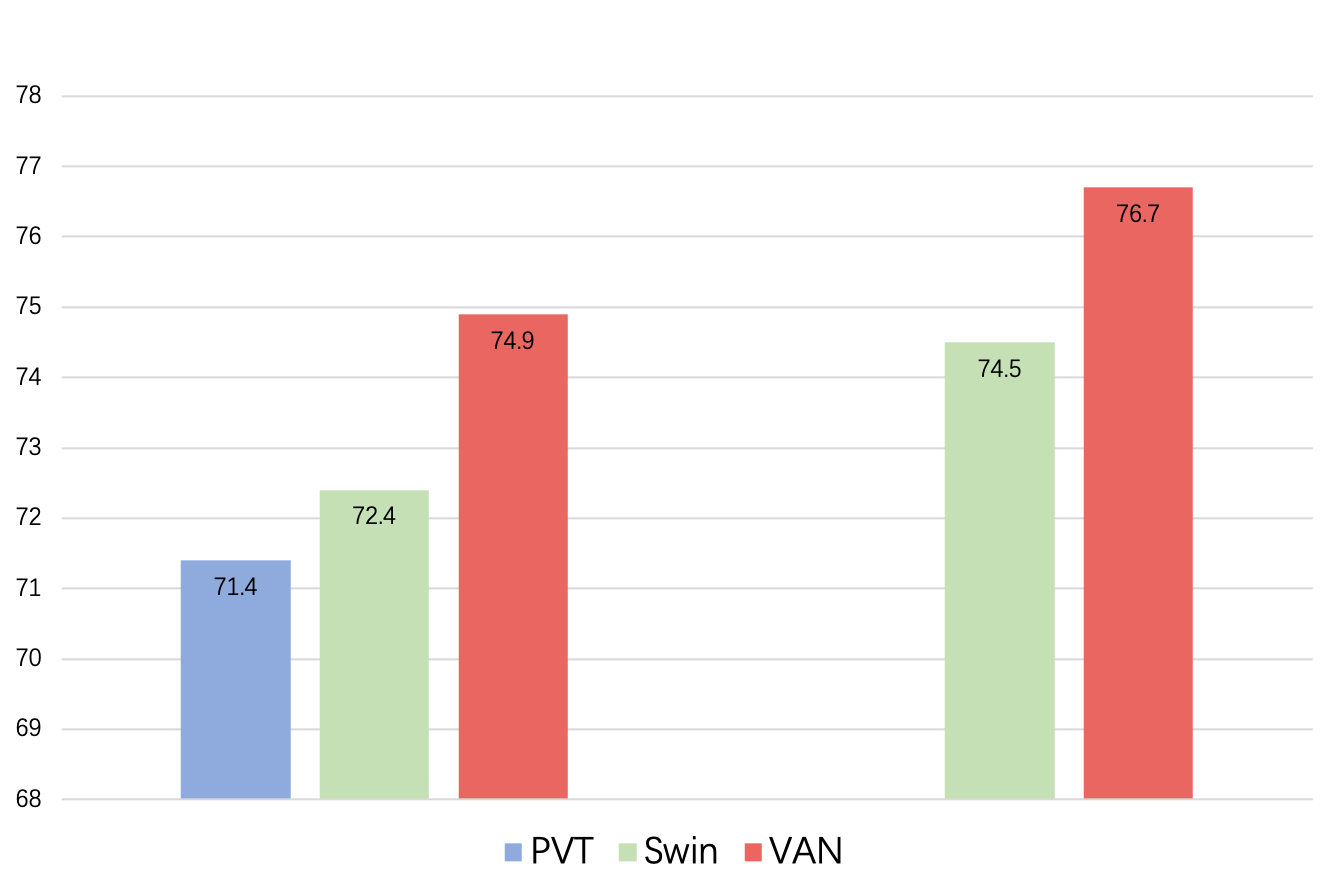}
    \put(5,62){Pose Estimation}
    \end{overpic}\hspace{1pt} 
    \\
    \caption{Comparing with similar level PVT~\cite{wang2021pyramid}, 
    Swin Transformer~\cite{liu2021swin} and ConvNeXt~\cite{liu2022convnet}
    on various tasks, including image classification, object detection, 
    semantic segmentation, instance segmentation and pose estimation.
    }\label{fig:compare_pvt_swin_convnext}
\end{figure*}

\begin{table}[t]
    \centering
    \setlength{\tabcolsep}{1.8mm}
    \caption{Ablation study of different kernel size $K$ in LKA. 
      Acc(\%) means Top-1 accuracy on ImageNet validation set. 
    }
    \begin{tabular}{l|c|c|c|c|c}
        \hline
        \textbf{Method} & $K$ & \textbf{Dilation} & \textbf{Params. (M)} & \textbf{GFLOPs}  & \textbf{Acc(\%)} \\
        \hline
        VAN-B0 & 7 & 2 &  4.03 &  0.85 &  74.8 \\
        VAN-B0 & 14 & 3 &  4.07  &  0.87 &  75.3 \\
        VAN-B0 & 21 & 3 & 4.11 &  0.88  &  75.4 \\
        VAN-B0 & 28 & 4 & 4.14 & 0.90 &  75.4 \\
        \hline
    \end{tabular}
    \label{Tab.ablation_kernelsize}
    \vspace{-2ex}
\end{table}

\begin{table}[!t]\centering
\renewcommand{\tabcolsep}{1.7mm}
\caption{Compare with the state-of-the-art methods on ImageNet validation set. Params means parameter. GFLOPs denotes floating point operations. Top-1 Acc represents Top-1 accuracy.FLOPs is }
\begin{tabular}{l|c|c|c}
    \hline
	Method & Params. (M) & GFLOPs & Top-1 Acc (\%)  \\
	\hline
	PVTv2-B0 ~\cite{wang2021pvtv2} & 3.4 & 0.6 & 70.5 \\
	T2T-ViT-7 ~\cite{Yuan_2021_ICCV} & 4.3 & 1.1 & 71.7 \\ 
	DeiT-Tiny/16~\cite{touvron2021training} & 5.7 & 1.3 & 72.2 \\
    TNT-Ti~\cite{tnt} & 6.1 & 1.4 & 73.9 \\
	\rowcolor{gray!25}{VAN-B0 } & 4.1 & 0.9 & \textbf{75.4} \\ 

	\hline
	ResNet18~\cite{he2016deep} & 11.7 & 1.8 & 69.8  \\
	PVT-Tiny~\cite{wang2021pyramid} & 13.2 & 1.9 &75.1 \\
	PoolFormer-S12~\cite{yu2021metaformer} & 11.9  & 2.0 & 77.2 \\   
	PVTv2-B1~\cite{wang2021pvtv2} & 13.1 & 2.1 & 78.7 \\
	\rowcolor{gray!25}{VAN-B1 } & 13.9 & 2.5 & \textbf{81.1} \\ 
	\hline
	ResNet50~\cite{he2016deep}   &25.6 &4.1 &76.5 \\
	ResNeXt50-32x4d~\cite{xie2017aggregated} &25.0 &4.3 &77.6  \\
	RegNetY-4G~\cite{regnet} & 21.0 & 4.0 & 80.0 \\
	DeiT-Small/16~\cite{touvron2021training}  & 22.1 & 4.6 & 79.8 \\
	T2T-ViT$_t$-14~\cite{Yuan_2021_ICCV} & 21.5 & 6.1 & 81.7 \\
	PVT-Small~\cite{wang2021pyramid}  & 24.5 & 3.8 & 79.8 \\
	TNT-S~\cite{tnt} & 23.8 & 5.2 & 81.3 \\
	ResMLP-24~\cite{touvron2021resmlp} & 30.0 & 6.0 & 79.4 \\
	gMLP-S~\cite{liu2021pay} & 20.0 & 4.5 & 79.6 \\
	Swin-T~\cite{liu2021swin} & 28.3 & 4.5 & 81.3 \\
	PoolFormer-S24~\cite{yu2021metaformer} & 21.4  & 3.6 & 80.3 \\   
	Twins-SVT-S~\cite{chu2021twins} & 24.0 & 2.8 & 81.7 \\
	PVTv2-B2~\cite{wang2021pvtv2} & 25.4 & 4.0 & 82.0 \\
	Focal-T~\cite{yang2021focal} & 29.1 & 4.9 & 82.2 \\
	ConvNeXt-T~\cite{liu2022convnet} & 28.6 & 4.5 & 82.1 \\  
	\rowcolor{gray!25}{VAN-B2 } & 26.6 & 5.0 & \textbf{82.8} \\
	\hline
	ResNet101~\cite{he2016deep}  &44.7 & 7.9 &77.4\\
	ResNeXt101-32x4d~\cite{xie2017aggregated} & 44.2 & 8.0 &78.8 \\
	Mixer-B/16~\cite{tolstikhin2021mlp} & 59.0 & 11.6 & 76.4 \\
	T2T-ViT$_t$-19~\cite{Yuan_2021_ICCV} & 39.2 & 9.8 & 82.4 \\
	PVT-Medium~\cite{wang2021pyramid} & 44.2 & 6.7 & 81.2\\
	Swin-S ~\cite{liu2021swin} & 49.6 & 8.7 & 83.0 \\
	ConvNeXt-S ~\cite{liu2021swin} & 50.1 & 8.7 & 83.1 \\
	PVTv2-B3 ~\cite{wang2021pvtv2} & 45.2 & 6.9 & 83.2 \\
	Focal-S~\cite{yang2021focal} & 51.1 & 9.1 & 83.5 \\
    \rowcolor{gray!25}{VAN-B3} & 44.8 & 9.0 & \textbf{83.9} \\
	\hline
	ResNet152~\cite{he2016deep} & 60.2 & 11.6 & 78.3 \\
	T2T-ViT$_t$-24~~\cite{Yuan_2021_ICCV} & 64.0 & 15.0 & 82.3 \\
	PVT-Large~~\cite{wang2021pyramid} & 61.4 & 9.8 & 81.7 \\
	TNT-B~~\cite{tnt} & 66.0 & 14.1 & 82.8 \\
	PVTv2-B4~\cite{wang2021pvtv2} & 62.6 & 10.1 & 83.6 \\
    \rowcolor{gray!25}{VAN-B4} & 60.3 & 12.2 & \textbf{84.2} \\
\hline
\end{tabular}
\label{tab:imagenet_cls}
\end{table}

\begin{table}[!t]\centering
\renewcommand{\tabcolsep}{1.7mm}
\caption{Compare with the state-of-the-art methods on ImageNet validation set. Params means parameter. GFLOPs denotes floating point operations. Top-1 Acc represents Top-1 accuracy.All models are pretrained on ImageNet-22K dataset.}
\begin{tabular}{l|c|c|c|c}
    \hline
	Method & Params. (M) & \makecell{Input \\ size}& GFLOPs & \makecell{Top-1 \\ Acc (\%)}  \\
	\hline
	Swin-S ~\cite{liu2021swin} & 50 & 224$^2$ & 8.7 & 83.2  \\
    ConvNeXt-S~\cite{liu2022convnet} & 50 & 224$^2$ & 8.7 & 84.6  \\
    \rowcolor{gray!25}	VAN-B4  & 60 & 224$^2$ & 12.2 & \textbf{85.7}  \\
    \hline
	ConvNeXt-S~\cite{liu2022convnet} & 50 & 384$^2$ & 25.5 & 85.8 \\
	\rowcolor{gray!25} VAN-B4  & 60 & 384$^2$ & 35.9 & \textbf{86.6} \\
    \hline
	Swin-B ~\cite{liu2021swin} & 88 & 224$^2$ & 15.4 & 85.2  \\
	ConvNeXt-B ~\cite{liu2022convnet} & 89 & 224$^2$ & 15.4 & 85.8 \\
    \rowcolor{gray!25}	VAN-B5  & 90 & 224$^2$ & 17.2 & \textbf{86.3}  \\
    \hline
	EffNetV2-L~\cite{tan2021efficientnetv2}  & 120 & 480$^2$ & 53.0 & 86.8 \\
    ViT-B/16~\cite{dosovitskiy2020image}  & 87 & 384$^2$ & 55.5 & 85.4 \\
	Swin-B ~\cite{liu2021swin} & 88 & 384$^2$ & 47.0 & 86.4 \\
	ConvNeXt-B ~\cite{liu2022convnet} & 89 & 384$^2$ & 45.1 & 86.8  \\
	\rowcolor{gray!25} VAN-B5  & 90 & 384$^2$ & 50.6 & \textbf{87.0} \\
	\hline
	Swin-L ~\cite{liu2021swin} & 197 & 224$^2$ & 34.5 & 86.3  \\
	ConvNeXt-L ~\cite{liu2022convnet} & 198 & 224$^2$ & 34.4 & 86.6  \\
	\rowcolor{gray!25}	VAN-B6  & 200 & 224$^2$ & 38.9 & \textbf{86.9}  \\
	\hline
    EffNetV2-XL~\cite{tan2021efficientnetv2} & 208 & 480$^2$ & 94.0 & 87.3  \\
    CoAtNet-3~\cite{dai2021coatnet} & 168 & 384$^2$ & 107.4 & 87.6  \\
	Swin-L ~\cite{liu2021swin} & 197 & 384$^2$ & 103.9 & 87.3 \\
	ConvNeXt-L ~\cite{liu2022convnet} & 198 & 384$^2$ & 101.0 & 87.5  \\
\rowcolor{gray!25}	VAN-B6  & 200 & 384$^2$ & 114.3 & \textbf{87.8} \\
\hline
\end{tabular}
\label{tab:22k_cls}
\end{table}

\begin{itemize}
    \item \textbf{DW-Conv.} DW-Conv can make use of the local contextual information of images. Without it, the classification performance will drop by 0.5\% (74.9\% vs. 75.4\%), showing the importance of local structural information in image processing.
    \item \textbf{DW-D-Conv.} DW-D-Conv denotes depth-wise dilation convolution which plays a role in capturing long-range dependence in LKA. Without it, the classification performance will drop by 1.3\% (74.1\% vs. 75.4\%) which confirms our viewpoint of long-range dependence is critical for visual tasks.
    \item \textbf{Attention Mechanism.} 
    The introduction of the attention mechanism can be regarded as
    making network achieve adaptive property. 
    Benefited from it, the VAN-B0 achieves about 1.1\% (74.3\% vs. 75.4\%) improvement.
    Besides, replacing attention with adding operation is also not 
    achieving a lower accuracy.
    \item \textbf{$1\times1$ Conv.} Here, $1\times1$ Conv captures relationship in channel dimension.  Combining with attention mechanism, it introduces adaptability in channel dimension. It brings about 0.8\% (74.6\% vs. 75.4\%) improvement which proves the necessity of the adaptability in channel dimension.
    \item \textbf{Sigmoid function.} Sigmoid function is a common 
    normalization function to normalize attention map from 0 to 1.
    However, we find it is not necessary for LKA module in our experiment.
    Without sigmoid, our VAN-B0 achieves 0.2\% (75.4\% vs. 75.2\%) improvement and 
    less computation.
\end{itemize}

Through the above analysis, we can find that our proposed LKA
can utilize local information, 
capture long-distance dependencies, 
and have adaptability in both channel and spatial dimension.
Furthermore, 
experimental results prove all properties are positive for recognition tasks.
Although standard convolution can make full use of the 
local contextual information, 
it ignores long-range dependencies and adaptability.
As for self-attention, although it can capture long-range dependencies and
has adaptability in spatial dimensions, 
it neglects the local information and 
the adaptability in the channel dimension.
Meanwhile, We also summarize above discussion in 
\tabref{Tab.operation_properties}.

Besides, we also conduct ablation study to decompose different size convolution kernels in ~\tabref{Tab.ablation_kernelsize}.
We can find that 
decomposing a 21$\times$21 convolution works better
than decomposing a 7$\times$7 convolution 
which demonstrates large kernel is critical for visual tasks. 
Decomposing a larger 28$\times$28 convolution, 
we find the gain is not obvious comparing with 
decomposing a 21$\times$21 convolution.
Thus, we choose to decompose a 21$\times$21 convolution by default.


\newcommand{\addImg}[1]{\includegraphics[width=.24\linewidth]{/cam_imgs/#1}}
\newcommand{\addImgs}[1]{\addImg{#1.jpg}  \addImg{#1_swin.jpg}  \addImg{#1_convnext.jpg}  \addImg{#1_van.jpg}}

\begin{figure}[t!]
  \centering
  \small
  \addImgs{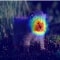} \\
  \addImgs{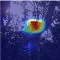} \\
  \addImgs{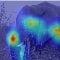} \\
  \addImgs{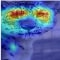} \\
  \addImgs{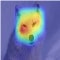} \\
  \quad input \qquad \qquad Swin-T \quad \qquad ConvNeXt-T \qquad VAN-B2
  \caption{Visualization results. 
    All images come from different categories in ImageNet validation set. 
    CAM is produced by using Grad-CAM~\cite{selvaraju2017grad}. 
    We compare different CAMs produced by  Swin-T~\cite{liu2021swin}, ConvNeXt-T~\cite{liu2022convnet} and VAN-B2.}
  \label{fig:cam}
\end{figure}

\myPara{Comparison with Existing Methods.}
\tabref{tab:imagenet_cls} presents the comparison of VAN with other MLPs, 
CNNs and ViTs.
VAN outperforms common CNNs (ResNet~\cite{he2016deep}, 
ResNeXt~\cite{xie2017aggregated}, ConvNeXt~\cite{liu2022convnet}, \etc.), 
ViTs (DeiT~\cite{touvron2021training}, PVT~\cite{wang2021pyramid} 
and Swin-Transformer~\cite{liu2021swin}, \etc.)
and MLPs (MLP-Mixer~\cite{tolstikhin2021mlp}, ResMLP~\cite{touvron2021resmlp}, 
gMLP~\cite{liu2021pay}, \etc.) 
with similar parameters and computational cost.   
We visually show the comparison of our method 
with similar level classical methods on different tasks
in ~\figref{fig:compare_pvt_swin_convnext},
which clearly reveals the improvement of our method.
In the following discussion, 
we will choose a representative network in each category. 

ConvNeXt~\cite{liu2022convnet} is a special CNN 
which absorbs the some advantages of ViTs such as large receptive field 
(7$\times$7 convolution)
and advanced training strategy(300 epochs, data augmentation, \etc).
Compared VAN with ConvNeXt~\cite{liu2022convnet}, 
VAN-B2 surpasses ConvNeXt-T by 0.7\% (82.8\% vs. 82.1\%) 
since VAN has larger receptive field and adaptive ability.
Swin-Transformer is a well-known ViT variant 
that adopts local attention and shifted window manner.
Due to that VAN
is friendly for 2D structural information, 
has larger receptive field and
achieves adaptability in channel dimension, 
VAN-B2 surpasses Swin-T by 1.5\% (82.8\% vs. 81.3\%).
As for MLPs, we choose gMLP~\cite{liu2021pay}.
VAN-B2 surpass gMLP-S~\cite{liu2021pay} by 3.2\% (82.8\% vs. 79.6\%)
which reflects the importance of locality.

\myPara{Throughput.} 
We test the throughput of the Swin transformer~\cite{liu2021swin} 
and VAN on some hardware environment with the RTX 3090.
Results are shown in~\tabref{tab_speed}.
Besides, we also plots the accuracy-throughput diagram on~\figref{fig:fps_result},
which clearly demonstrates 
VAN achieves a better accuracy-throughput
trade-off than swin transformer~\cite{liu2021swin}.

\subsubsection{Visualization} 
Class activation mapping (CAM) is a popular tool to 
visualize the discriminative regions (attention maps).
We adopt Grad-CAM~\cite{selvaraju2017grad} to visualize the attentions
on the ImageNet validation set produced by VAN-B2 model. 
Results in \figref{fig:cam} show that VAN-B2 can 
clearly focus on the target objects. 
Thus, the visualizations intuitively demonstrate 
the effectiveness of our method. 
Furthermore, we also compare different CAM produced by Swin-T~\cite{liu2021swin}, ConvNeXt-T~\cite{liu2022convnet} and VAN-B2. 
We can find that the activation area of VAN-B2 is more accurate. 
In particular, our method shows obvious advantages when the object is dominant in an image (last 3 lines in~\figref{fig:cam}),
which demonstrates its ability to capture long-range dependence.

\subsubsection{Pretraining on ImageNet-22K.}

\myPara{Settings.} 
ImageNet-22K is a large-scale image
classification dataset, which contains about 14M images and 
21841 categories. 
Following swin transformer~\cite{liu2021swin} and ConvNeXt~\cite{liu2022convnet},
we use it to pretrain our VAN for 90 epochs without EMA.
The batch size is set as 8,196.
Other training details are same with ImageNet-1K settings.
After pretrained on ImageNet-22K, we fine-tune our
model on ImageNet-1K for 30 epochs. 
We pretrain our model with 224 $\times$ 224 input 
and fine-tune our model with 224 $\times$ 224
and  384 $\times$ 384 respectively.

\myPara{Results.}
We compare current state-of-the-art CNNs(\eg{ConvNeXt~\cite{liu2022convnet}, EFFNetV2~\cite{tan2021efficientnetv2}}) 
and ViTs(\eg{Swin Transformer~\cite{liu2021swin}}, ViT~\cite{dosovitskiy2020image}
and CoAtNet~\cite{dai2021coatnet}).
As shown in~\tabref{tab:22k_cls}, VAN achieves 87.8\% Top-1 accuracy with 200M parameters
and surpasses the same level ViT~\cite{dosovitskiy2020image}, Swin Transformer~\cite{liu2021swin}, EFFNetV2~\cite{tan2021efficientnetv2} and ConvNeXt~\cite{liu2022convnet}
on different resolution,
which proves the strong capability to adapt large-scale pretraining.

\begin{table}[t]
  \centering
  \setlength{\tabcolsep}{1mm}
\caption{Object detection on COCO 2017 dataset. \#P means parameter. 
RetinaNet 1$\times$ denotes models are based on RetinaNet~\cite{lin2017focal} and 
we train them for 12 epochs.}
\begin{tabular}{l|c|lcc|lcc}
\hline
\multirow{2}{*}{Backbone} &\multicolumn{7}{c}{RetinaNet 1$\times$} \\
\cline{2-8} 
& \#P (M) &AP &AP$_{50}$ &AP$_{75}$ &AP$_S$ &AP$_M$ &AP$_L$  \\
\hline 
\rowcolor{gray!25} VAN-B0  & {13.4} &  \textbf{38.8} &  \textbf{58.8} & \textbf{41.3} &   \textbf{23.4} &  \textbf{42.8} &  \textbf{50.9} \\
\hline
ResNet18~\cite{he2016deep} & {21.3} & 31.8 & 49.6 & 33.6 & 16.3 & 34.3 & 43.2 \\
PoolFormer-S12~\cite{wang2021pyramid} &21.7& {36.2}& {56.2}& {38.2}& {20.8}& {39.1} & {48.0} \\
PVT-Tiny~\cite{wang2021pyramid} &23.0& {36.7}& {56.9}& {38.9}& {22.6}& {38.8} & {50.0} \\
\rowcolor{gray!25} VAN-B1  & {23.6} &  \textbf{42.3} &  \textbf{63.1} & \textbf{45.1} &  \textbf{26.1} &  \textbf{46.2} &  \textbf{54.1} \\
\hline
ResNet50~\cite{he2016deep} &37.7 & 36.3 & 55.3 & 38.6 & 19.3 & 40.0 & 48.8 \\
PVT-Small~\cite{wang2021pyramid} & {34.2} & {40.4} & {61.3} & {43.0} & {25.0} & {42.9} & {55.7} \\
PoolFormer-S24~\cite{yu2021metaformer} & {31.1} & {38.9} & {59.7} & {41.3} & {23.3} & {42.1} & {51.8} \\
PoolFormer-S36~\cite{yu2021metaformer} & {40.6} & {39.5} & {60.5} & {41.8} & {22.5} & {42.9} & {52.4} \\
\rowcolor{gray!25} VAN-B2  & {36.3} &  \textbf{44.9} &  \textbf{65.7} & \textbf{48.4} &  \textbf{27.4} &  \textbf{49.2} &  \textbf{58.7} \\
\hline
ResNet101~\cite{he2016deep} &56.7  & 38.5 & 57.8 & 41.2 & 21.4 & 42.6 & 51.1 \\
PVT-Medium~\cite{wang2021pyramid} & {53.9} & {41.9} & {63.1} & {44.3} & {25.0} & {44.9} & {57.6} \\
\rowcolor{gray!25} VAN-B3   & 54.5 & \textbf{47.5} & \textbf{68.4} & \textbf{51.2} & \textbf{30.9} & \textbf{52.1} & \textbf{62.4}  \\
\hline
\end{tabular}
\label{tab:retina_1x}
\end{table}

\begin{table}[!t]\centering
  \setlength{\tabcolsep}{.7mm}
\caption{Object detection and instance segmentation on COCO 2017 dataset. 
\#P means parameter. 
Mask R-CNN 1$\times$ denotes models are based on Mask R-CNN~\cite{He_maskrcnn} and we train them for 12 epochs.
AP$^{b}$ and AP$^{m}$ refer to bounding box AP and mask AP respectively.
}
\begin{tabular}{l|c|lcc|lcc}
\hline
\multirow{2}{*}{Backbone} &\multicolumn{7}{c}{Mask R-CNN 1$\times$} \\
\cline{2-8} 
& \#P (M) & AP$^{b}$ &AP$^{b}_{50}$ &AP$^{b}_{75}$ &AP$^{m}$ &AP$^{m}_{50}$ &AP$^{m}_{75}$  \\
\hline
\rowcolor{gray!25} VAN-B0  & {23.9} &  \textbf{40.2} &  \textbf{62.6} & \textbf{44.4} &  \textbf{37.6} &  \textbf{59.6} &  \textbf{40.4} \\
\hline
ResNet18~\cite{he2016deep} & {31.2} & 34.0 & 54.0 & 36.7 & 31.2 & 51.0 & 32.7\\ 
PoolFormer-S12~\cite{yu2021metaformer} & {31.6} & 37.3 & 59.0 & 40.1 & 34.6 & 55.8 & 36.9\\ 
PVT-Tiny~\cite{wang2021pyramid} &32.9& {36.7}& {59.2}& {39.3}& {35.1}& {56.7} & {37.3} \\
\rowcolor{gray!25} VAN-B1  & {33.5} &  \textbf{42.6} &  \textbf{64.2} & \textbf{46.7} &  \textbf{38.9} &  \textbf{61.2} &  \textbf{41.7} \\
\hline
ResNet50~\cite{he2016deep} & 44.2& 38.0 & 58.6 & 41.4 & 34.4 & 55.1 & 36.7\\
PVT-Small~\cite{wang2021pyramid}  & {44.1} &{40.4} & {62.9} & {43.8} & {37.8} & {60.1} & {40.3}\\
PoolFormer-S24~\cite{yu2021metaformer}  & {41.0} &{40.1} & {62.2} & {43.4} & {37.0} & {59.1} & {39.6}\\
PoolFormer-S36~\cite{yu2021metaformer}  & {50.5} &{41.0} & {63.1} & {44.8} & {37.7} & {60.1} & {40.0}\\
\rowcolor{gray!25} VAN-B2  & {46.2} &  \textbf{46.4} &  \textbf{67.8} & \textbf{51.0} &  \textbf{41.8} &  \textbf{65.2} & \textbf {44.9} \\
\hline
ResNet101~\cite{he2016deep} &63.2 & 40.4 & 61.1 & 44.2 & 36.4 & 57.7 & 38.8 \\
ResNeXt101-32x4d~\cite{xie2017aggregated} & {62.8} & 41.9 & 62.5 & {45.9} & 37.5 & 59.4 & 40.2 \\
PVT-Medium~\cite{wang2021pyramid} &63.9 & {42.0} &{64.4} &45.6 &{39.0}& {61.6}& {42.1}\\
\rowcolor{gray!25} VAN-B3    & 64.4 &  \textbf{48.3} &  \textbf{69.6} &  \textbf{53.3} & \textbf{43.4} &  \textbf{67.0} & \textbf{46.8} \\
\hline
\end{tabular}
\label{tab:maskrcnn_1x}
\end{table}

\begin{table}[!t]
  \centering
  \setlength{\tabcolsep}{.2mm}
\caption{Comparison with the state-of-the-art vision backbones on COCO 2017 benchmark. All models are trained for 36 epochs. We calculate FLOPs with input size of 1,280 $\times$ 800.}
\begin{tabular}{l|c|ccc|ccc}
\hline
Backbone & \quad Method \quad  & AP$^\text{b}$ & AP$^\text{b}_\text{50}$ & AP$^\text{b}_\text{75}$ & \#P (M) & GFLOPs \\
 \hline
Swin-T~\cite{he2016deep}& \multirow{3}{*}{\makecell{Mask R-CNN~\cite{He_maskrcnn}}}& 46.0 & 68.1 & 50.3 & 48 & 264  \\
ConvNeXt-T~\cite{liu2021swin} & & 46.2 & 67.9 & 50.8 & 48 & 262  \\
\rowcolor{gray!25} VAN-B2  & & \textbf{48.8} & \textbf{70.0} & \textbf{53.6} & 46 & 273  \\
\hline
 ResNet50~\cite{he2016deep} & \multirow{4}{*}{\makecell{Cascade\\Mask R-CNN~\cite{cai2019cascade}}} & 46.3 & 64.3 & 50.5 & 82 & 739  \\
 Swin-T~\cite{liu2021swin} & & 50.5 & 69.3 & 54.9 & 86 & 745  \\
ConvNeXt-T ~\cite{liu2022convnet} & & {50.4} & {69.1} & {54.8} & 86 &  741 \\
 \rowcolor{gray!25} VAN-B2  & & \textbf{52.0} & \textbf{70.9} & \textbf{56.4} & 84 & 752 \\
\hline
ResNet50~\cite{he2016deep} & \multirow{3}{*}{ATSS~\cite{zhang2020bridging}} & 43.5	& 61.9 & 47.0 & 32 & 205  \\
Swin-T~\cite{liu2021swin} & & 47.2 & 66.5 & 51.3 & 36 & 215  \\
 \rowcolor{gray!25} VAN-B2  &  & \textbf{50.2} & \textbf{69.3} & \textbf{55.1} & 34 & 221  \\
\hline
ResNet50~\cite{he2016deep} & \multirow{3}{*}{GFL~\cite{li2020generalized}} & 44.5 & 63.0 & 48.3 &32 & 208  \\
Swin-T~\cite{liu2021swin} & & 47.6 &66.8& 51.7 &36 & 215 \\	
 \rowcolor{gray!25} VAN-B2  & & \textbf{50.8} & \textbf{69.8} & \textbf{55.7} & 34 & 224  \\

\hline
\end{tabular}
\label{tab:det_sota}
\end{table}

\begin{table}[t]
\centering
\setlength{\tabcolsep}{1mm}
\caption{Results of semantic segmentation on ADE20K~\cite{zhou2019semantic} validation set. The upper and lower part are obtained under two different training/validation schemes following ~\cite{yu2021metaformer} and~\cite{liu2021swin}. We calculate FLOPs with input size 512 $\times$ 512 for Semantic FPN~\cite{kirillov2019panoptic} and 2,048 $\times$ 512 for UperNet~\cite{xiao2018unified}.}
\label{tab:seg}
\begin{tabular}{c|c|c|c|c}
\hline
Method & Backbone & \#P(M) & GFLOPs & mIoU (\%)  \\
\hline
& PVTv2-B0~\cite{wang2021pvtv2} & 8 & 25 & 37.2 \\
&  \cellcolor{gray!25} VAN-B0 &   \cellcolor{gray!25} 8 &  \cellcolor{gray!25} 26 &  \cellcolor{gray!25} \textbf{38.5} \\
\cline{2-5}
& ResNet18~\cite{he2016deep} & 16 & 32 & 32.9 \\
& PVT-Tiny~\cite{wang2021pyramid} & 17 & 33 & 35.7 \\
& PoolFormer-S12~\cite{yu2021metaformer} & 16 & 31 & 37.2 \\
& PVTv2-B1~\cite{wang2021pvtv2} & 18 & 34 & 42.5 \\
& \cellcolor{gray!25} VAN-B1 & \cellcolor{gray!25} 18 & \cellcolor{gray!25} 35 & \cellcolor{gray!25} \textbf{42.9} \\
\cline{2-5}
& ResNet50~\cite{he2016deep} & 29 & 46 & 36.7\\
Semantic & PVT-Small~\cite{wang2021pyramid} & 28 & 45 & 39.8 \\
FPN \cite{kirillov2019panoptic} & PoolFormer-S24~\cite{yu2021metaformer} & 23 & 39 & 40.3 \\
& PVTv2-B2~\cite{wang2021pvtv2} & 29 & 46  & 45.2 \\
& \cellcolor{gray!25} VAN-B2 & \cellcolor{gray!25} 30 & \cellcolor{gray!25} 48 & \cellcolor{gray!25} \textbf{46.7} \\
\cline{2-5}
& ResNet101~\cite{he2016deep} & 48 & 65 & 38.8\\
& ResNeXt101-32x4d~\cite{xie2017aggregated} & 47 & 65 & 39.7 \\
& PVT-Medium~\cite{wang2021pyramid} & 48 & 61 & 43.5\\
& PoolFormer-S36~\cite{yu2021metaformer} & 35 & 48 & 42.0 \\
& PVTv2-B3~\cite{wang2021pvtv2} & 49 & 62 & 47.3 \\
& \cellcolor{gray!25} VAN-B3 & \cellcolor{gray!25} 49 & \cellcolor{gray!25} 68 & \cellcolor{gray!25} \textbf{48.1} \\

\hline
\hline

UperNet~\cite{xiao2018unified} & \multirow{3}{*}{ResNet-101~\cite{he2016deep}} & 86 & 1029 & 44.9  \\
OCRNet~\cite{yuan2020object} &  & 56 & 923 & 45.3 \\
HamNet~\cite{geng2021attention} & & 69 & 1111 & 46.8  \\

\hline
\multirow{9}{*}{UperNet~\cite{xiao2018unified}}  
& Swin-T~\cite{liu2021swin}  & 60 & 945 & 46.1 \\
& ConvNeXt-T~\cite{liu2022convnet}  & 60 & 939 & 46.7 \\
& \cellcolor{gray!25} VAN-B2 & \cellcolor{gray!25} 57 
& \cellcolor{gray!25} 948 & \cellcolor{gray!25} \textbf{50.1} \\
& Swin-S~\cite{liu2021swin}  & 81 & 1038 & 49.3  \\
& ConvNeXt-S~\cite{liu2022convnet}  & 82 & 1027 & 49.5  \\
& \cellcolor{gray!25} VAN-B3 & \cellcolor{gray!25}  75 
& \cellcolor{gray!25} 1030 & \cellcolor{gray!25} \textbf{50.6} \\
& Swin-B~\cite{liu2021swin}  & 121 & 1188 & 49.7  \\
& ConvNeXt-B~\cite{liu2022convnet}  & 122 & 1170 & 49.9  \\
& \cellcolor{gray!25} VAN-B4 & \cellcolor{gray!25} 90 &
\cellcolor{gray!25}1098 & \cellcolor{gray!25} \textbf{52.2} \\

\hline
\end{tabular}
\end{table}
\vspace{-2mm}

\begin{table}[!t]\centering
\renewcommand{\tabcolsep}{1.7mm}
\caption{Compare with the state-of-the-art methods on ADE20K validation set. Params means parameter. GFLOPs denotes floating point operations.All models are pretrained on ImageNet-22K dataset.
We calculate FLOPs with input size 2560 $\times$ 640 for 640 input image and 
2048 $\times$ 512 for 512 input image.}
\begin{tabular}{l|c|c|c|c}
    \hline
	Method & Params. (M) & \makecell{Input \\ size}& GFLOPs & \makecell{mIoU}  \\
	\hline
	Swin-B ~\cite{liu2021swin} & 121 & 640$^2$ & 1841 & 51.7  \\
	ConvNeXt-B ~\cite{liu2022convnet} & 122 & 640$^2$ & 1828 & 53.1 \\
    \rowcolor{gray!25}	VAN-B5  & 117 & 512$^2$ & 1208 & \textbf{53.9}  \\
	\hline
	Swin-L ~\cite{liu2021swin} & 234 & 640$^2$ & 2468 & 53.5 \\
	ConvNeXt-L ~\cite{liu2022convnet} & 235 & 640$^2$ & 2458 & 53.7  \\
    \rowcolor{gray!25}	VAN-B6  & 231 & 512$^2$ & 1658 & \textbf{54.7} \\
\hline
\end{tabular}
\label{tab:22k_seg}
\end{table}

\subsection{Object Detection}
\myPara{Settings.} We conduct object detection and instance segmentation experiments on COCO 2017 benchmark~\cite{lin2014microsoft},
which contains 118K images in training set and 5K images in validation set.  
MMDetection~\cite{chen2019mmdetection} is used as the codebase to implement detection models.  
For fair comparison, we adopt the same training/validating strategies with Swin Transformer~\cite{liu2021swin} and PoolFormer~\cite{yu2021metaformer}. 
Many kinds of detection models 
(\eg{Mask R-CNN~\cite{He_maskrcnn}, RetinaNet~\cite{lin2017focal}, Cascade Mask R-CNN~\cite{cai2019cascade}, Sparse R-CNN~\cite{sun2021sparse} ,\etc.}) 
are included to demonstrate the effectiveness of our method.
All backbone models are pre-trained on ImageNet training set.

\myPara{Results.} According to ~\tabref{tab:retina_1x} and ~\tabref{tab:maskrcnn_1x}, we find that VAN surpasses
CNN-based method ResNet~\cite{he2016deep} and 
transformer-based method PVT~\cite{wang2021pyramid} 
with a large margin
under RetinaNet~\cite{lin2017focal} 1x and Mask R-CNN~\cite{He_maskrcnn} 1x settings. 
Besides, we also compare the state-of-the-art methods Swin transformer~\cite{liu2021swin} and ConvNeXt~\cite{liu2022convnet} in ~\tabref{tab:det_sota}. 
Results show that VAN achieves the state-of-the-art performance with different detection methods such as Mask R-CNN~\cite{He_maskrcnn} and Cascade Mask R-CNN~\cite{cai2019cascade}. 

\subsection{Semantic Segmentation}

\myPara{Settings.} We conduct experiments on ADE20K~\cite{zhou2019semantic},
which contains 150 semantic categories for semantic segmentation.
It consists of 20,000, 2,000 and 3,000 separately for training, validation and testing.
MMSEG~\cite{mmseg2020} is used as the base framework and two famous segmentation heads, Semantic FPN~\cite{kirillov2019panoptic} and UperNet~\cite{xiao2018unified},
are employed for evaluating our VAN backbones.
For a fair comparison, we adopt two training/validating schemes following~\cite{yu2021metaformer} and~\cite{liu2021swin} 
and quantitative results on the validation set are shown in the upper and lower part in~\tabref{tab:seg}, respectively. 
All backbone models are pre-trained on ImageNet-1K or
ImageNet-22K training set.

\myPara{Results.}
From the upper part in~\tabref{tab:seg}, compared with different backbones using FPN~\cite{kirillov2019panoptic},
VAN-Based methods are superior to CNN-based (ResNet~\cite{he2016deep}, ResNeXt~\cite{xie2017aggregated}) or transformer-based (PVT~\cite{wang2021pyramid}, PoolFormer~\cite{yu2021metaformer}, PVTv2~\cite{wang2021pvtv2}) methods.
For instance, we surpass four PVTv2~\cite{wang2021pvtv2} variants by +1.3 (B0), +0.4 (B1), +1.5 (B2), +0.8 (B3) mIoU under comparable parameters and FLOPs.
In the lower part in~\tabref{tab:seg}, when compared with previous state-of-the-art CNN-based methods and Swin-Transformer based methods, 
four VAN variants also show excellent performance with comparable parameters and FLOPs.
For instance, based on UperNet~\cite{xiao2018unified}, VAN-B2 is +5.2 and +4.0 mIoU higher than ResNet-101 and Swin-T, respectively.
For ImageNet-22K pretrained models, VAN also performs better than 
Swin transformer~\cite{liu2021swin} and ConvNeXt~\cite{liu2022convnet}
with less computational overhead, 
which is shown in~\tabref{tab:22k_seg}.


\subsection{Panoptic Segmentation}

\begin{table}[!t]\centering
\renewcommand{\tabcolsep}{1.7mm}
\caption{Experimental results on COCO panoptic segmentation.
$^{*}$ means model is pretrained on ImageNet-22K dataset.
All methods are based on Mask2Former~\cite{cheng2022masked}.
PQ means panoptic quality.}
\begin{tabular}{l|c|c|c|c|c}
    \hline
	Backbone & Query type & Epochs & PQ & PQ$^{Th}$ & PQ$^{St}$  \\
	\hline
	Swin-T & 100 queries & 50 & 53.2  & 59.3 & 44.0  \\
\rowcolor{gray!25}	VAN-B2 & 100 queries & 50 & 54.9  & 61.2 & 45.3 \\
	Swin-L$^{*}$ & 200 queries & 50 & 57.8 & 64.2 & 48.1  \\
\rowcolor{gray!25}	VAN-B6$^{*}$ & 200 queries & 50 & \textbf{58.2}  & \textbf{64.8} & \textbf{48.2} \\
\hline
\end{tabular}
\label{tab:pano_seg}
\end{table}

\myPara{Settings.}
We conduct our panoptic segmentation on COCO panoptic segmentation dataset~\cite{lin2014microsoft}
and choose Mask2Former~\cite{cheng2022masked} as our segmentation head.
For fair comparison, we adopt the default settings in MMDetection~\cite{chen2019mmdetection} and same training/validating scheme 
as Mask2Former~\cite{cheng2022masked}. 
All backbone models are pre-trained on ImageNet-1K or ImageNet-22K set.

\myPara{Results.}
As shown in~\tabref{tab:pano_seg}, we 
observe that VAN outperforms Swin transformer 
for both large and small models.
Here, VAN-B2 exceeds Swin-T +1.7 PQ.
Besides, it is worth noting that
VAN-B6 achieves 58.2 PQ, which 
set new state-of-the-art performance 
for panoptic segmentation task.

\subsection{Pose Estimation}
\myPara{Settings.} 
We conduct pose estimation experiments on COCO human pose estimation dataset,
which contains 200K images with 17 keypoints.
Models are trained on COCO train 2017 set and tested on COCO val 2017 set.
We adopt SimpleBaseline~\cite{xiao2018simple} as our decoder part,
which is same with Swin transformer~\cite{liu2021swin} and PVT~\cite{wang2021pyramid}.
All experiments are based on MMPose~\cite{mmpose2020}.

\myPara{Results.}
Experimental results are shown on~\tabref{tab:pose_sota}.
For 256 $\times$ 192 input, 
VAN-B2 outperform Swin-T and PVT-S~\cite{wang2021pyramid} 
2.5AP (74.9 vs. 72.4) and 3.5AP (74.9 vs. 71.4) and  with simliar 
computing and parameters.
Furthermore, VAN-B2 exceeds Swin-B 2AP (74.9 vs. 72.9) and 1.8AP (76.7 vs. 74.9)
for 256 $\times$ 192 and 384 $\times$ 288 respectively with less
computation and parameters.
In addition to transformer-based models, VAN-B2 also surpasses
popular CNN-based 
model HRNet-W32~\cite{wang2020deep}.

\begin{table}[!t]
  \centering
  \setlength{\tabcolsep}{.2mm}
\caption{Comparison with the state-of-the-art vision backbones on COCO benchmark
for pose estimation. 
Models are based SimpleBaseline~\cite{xiao2018simple}.}
\begin{tabular}{l|c|cccccc}
\hline
Backbone & Input size & AP & AP$^\text{50}$ & AP$^\text{75}$ & AR & \#P (M) & GFLOPs \\
\hline
HRNet-W32~\cite{wang2020deep} & 256 $\times$ 192 & 74.4 & 90.5 & 81.9 & 78.9 & 28.5 & 7.1  \\
PVT-S~\cite{wang2021pyramid}& 256 $\times$ 192 & 71.4 & 89.6 & 79.4 & 77.3 & 28.2 & 4.1  \\
Swin-T~\cite{liu2021swin} & 256 $\times$ 192 & 72.4 & 90.1 & 80.6 & 78.2 & 32.8 & 6.1  \\
Swin-B~\cite{liu2021swin} & 256 $\times$ 192 & 72.9 & 89.9 & 80.8 & 78.6 & 93.2 & 18.6  \\
\rowcolor{gray!25} VAN-B2  & 256 $\times$ 192 & \textbf{74.9} & \textbf{90.8} & \textbf{82.5} & \textbf{80.3} & 30.3 & 6.1  \\
\hline
HRNet-W32~\cite{wang2020deep} & 384 $\times$ 288 & 75.8 & 90.6 & 82.7 & 81.0 & 28.5 & 16.0  \\
Swin-B~\cite{liu2021swin} & 384 $\times$ 288 & 74.9 & 90.5 & 81.8 & 80.3 & 93.2 & 39.2  \\
\rowcolor{gray!25} VAN-B2~\cite{liu2021swin} & 384 $\times$ 288 & \textbf{76.7} & \textbf{91.0} &\textbf{ 83.1} & \textbf{81.7} & 30.3 & 13.6  \\
\hline
\end{tabular}
\label{tab:pose_sota}
\end{table}

\subsection{Fine-grain Classification}
We conduct fine-grain classification on CUB-200 dataset~\cite{welinder2010caltech},
which is a common fine-grain classification benchmark
and contains 11,788 images of 200 subcategories belonging to birds.
We do not design specific algorithm for this task and only 
replace the last linear layer for 200 categories.
We implement our model based on mmclassification~\cite{2020mmclassification}.
Results on~\tabref{tab:fine_grain_cls} show that
VAN-B4 achieves 91.3 \% Top-1 accuracy without 
any specially designed Algorithms,
which exceeds DeiT~\cite{touvron2021training}
and ViT-B~\cite{dosovitskiy2020image}.


\begin{table}[!t]\centering
\renewcommand{\tabcolsep}{1.7mm}
\caption{Experimental results on CUB-200 fine-grain classification dataset.
$^{*}$ means model is pretrained on ImageNet-22K dataset.}
\begin{tabular}{l|c|c}
    \hline
	Method & Backbone & Top-1 Acc (\%)  \\
	\hline
	ResNet-50~\cite{he2016deep} & ResNet-101 &  84.5  \\
	ViT~\cite{dosovitskiy2020image}  & ViT-B\_16$^{*}$ & 90.3  \\
	DeiT~\cite{touvron2021training}  & DeiT-B$^{*}$ & 90.0  \\
\rowcolor{gray!25}	VAN  & VAN-B4$^{*}$ & 91.3  \\
\hline
\end{tabular}
\label{tab:fine_grain_cls}
\end{table}

\subsection{Saliency Detection}
We conduct saliency detection base on EDN~\cite{wu2022edn}.
We replace the backbone with VAN and 
hold experiments on common saliency detection
benchmarks, including DUTS~\cite{wang2017learning}.
DUT-O~\cite{yang2013saliency} and PASCAL-S~\cite{li2014secrets}.
Results on~\tabref{tab:saliency} show that
VAN clearly surpasses other backbones
ResNet~\cite{he2016deep} and PVT~\cite{wang2021pyramid}
on all datasets.

\begin{table}[!t]
  \centering
  \setlength{\tabcolsep}{.2mm}
\caption{Comparing with different backbones on saliency detection task.}
\begin{tabular}{l|cc|cc|cc}
\hline
\multirow{2}{*}{Backbone} &\multicolumn{2}{c}{DUTS-TE} 
& \multicolumn{2}{c}{DUT-O}
& \multicolumn{2}{c}{PASCAL-S}  \\ 
 & $F_{max}$ & MAE & $F_{max}$ & MAE & $F_{max}$ & MAE \\
\hline
ResNet18~\cite{he2016deep}& 0.853 & 0.044 & 0.769 & 0.056 & 0.854 & 0.071   \\
PVT-T~\cite{wang2021pyramid} & 0.876 & 0.039 & 0.813 & 0.052 & 0.868  & 0.067  \\
\rowcolor{gray!25} VAN-B1  & \textbf{0.912} & \textbf{0.030} & \textbf{0.835} & \textbf{0.046} & \textbf{0.893} & \textbf{0.055} \\
\hline
ResNet50~\cite{he2016deep}& 0.873 & 0.038 & 0.786 & 0.051 & 0.864 & 0.065   \\
PVT-S~\cite{wang2021pyramid} & 0.900 & 0.032 & 0.832 & 0.050 & 0.883 & 0.060  \\
\rowcolor{gray!25} VAN-B2  & \textbf{0.919} & \textbf{0.028} & \textbf{0.844} & \textbf{0.045} & \textbf{0.897} & \textbf{0.053}  \\
\hline
\end{tabular}
\label{tab:saliency}
\end{table}

\section{Discussion }

Recently, transformer-based models quickly conquer various vision leaderboards.
As we know that self-attention is just a special attention mechanism.
However, people gradually adopt self-attention by default
and ignore underlying attention methods.
This paper proposes a novel attention module LKA and 
CNN-based network VAN. which surpasses 
\sArt transformer-based methods for vision tasks.
We hope this paper can promote people to rethink 
whether self-attention is irreplaceable and
which kind of attention is more suitable for visual tasks.

\section{Future Work}
\label{sec:future_work}
In the future, we will continue perfecting VAN in followings directions:

\begin{itemize}
  \item \textbf{Continuous improvement of the structure itself.} In this paper, we only demonstrate an intuitive structure. 
  There are a lot of potential improvements such as adopting different kernel size, introducing multi-scale structure~\cite{pami21Res2net} and using multi-branch structure~\cite{szegedy2015going}. 
 
  \item \textbf{Large-scale self-supervised learning and transfer learning.} 
  VAN naturally combines the advantages of CNNs and ViTs.  
    On the one hand, VAN can make use of the 2D structure information of images.
    On the other hand, VAN can dynamically adjust the output 
    according to the input image which is suit for self-supervised learning and transfer learning~\cite{bao2021beit,he2021masked}.
    Combining the above two points, we believe VAN can achieve better performance in 
    image self-supervised learning and transfer learning field.
    
  \item \textbf{More application areas.} Due to the limited resource, 
  we only show excellent performance in visual tasks.
  Whether VANs can perform well in other areas like TCN~\cite{bai2018empirical} in NLP
  is still worth exploring. 
  we look forward to seeing VANs 
  becoming a general model. 
  
\end{itemize}

\section{Conclusion}

In this paper, we present a novel visual attention LKA 
which combines the advantages of convolution and self-attention. 
Based on LKA, we build a vision backbone VAN 
that achieves the state-of-the-art performance in some visual tasks, 
including image classification, object detection, semantic segmentation, \etc.
In the future, we will continue to improve this framework 
from the directions mentioned in ~\secref{sec:future_work}.


%



\section*{Acknowledgments}

This work was supported by the National Key R\&D Program(2018AAA0100400), 
the Natural Science Foundation of China (Project
61521002, 61922046), and Tsinghua-Tencent Joint Laboratory for Internet Innovation Technology.

\bibliographystyle{IEEEtran}
\bibliography{VAN.bib}

\end{document}